\documentclass[10pt,twocolumn,letterpaper]{article}

\newif\ifarxiv
\arxivtrue

\ifarxiv
\usepackage[pagenumbers, algorithms]{wacv}
\else
\usepackage[algorithms]{wacv}
\fi

\usepackage{graphicx}
\usepackage{booktabs}
\usepackage{appendix}
\usepackage{amsmath}
\usepackage{amssymb}
\usepackage{enumitem}
\setlist[itemize]{label=\textbullet}
\usepackage{etoolbox}
\usepackage{subcaption}
\usepackage[dvipsnames]{xcolor}
\usepackage[export]{adjustbox}
\usepackage{placeins}
\usepackage{longtable}
\usepackage{stix}
\usepackage{bibentry}
\usepackage{fancyhdr}
\usepackage{sectsty}
\usepackage{algorithm}
\usepackage{algpseudocode}
\usepackage{titlesec}

\titlespacing*{\section}
{0pt}{1.5ex plus 1ex minus .2ex}{0.5ex plus .2ex}
\usepackage[accsupp]{axessibility}

\usepackage[pagebackref,breaklinks,colorlinks]{hyperref}

\usepackage[capitalize]{cleveref}
\crefname{figure}{Fig.}{Figs.}
\Crefname{figure}{Fig.}{Figs.}

\crefname{table}{Tab.}{Tabs.}
\Crefname{table}{Tab.}{Tabs.}

\crefname{section}{Sec.}{Secs.}
\Crefname{section}{Sec.}{Secs.}

\crefname{subsection}{Sec.}{Secs.}
\Crefname{subsection}{Sec.}{Secs.}

\newcommand*{\simsym}{\mathord\sim}

\def\wacvPaperID{2118}
\def\confName{WACV}
\def\confYear{2025}

\begin{document}

\title{A Rapid Test for Accuracy and Bias of Face Recognition Technology}

\author{
Manuel Knott$^{1*}$
~~
Ignacio Serna$^{1,2*}$
~~
Ethan Mann$^{1*}$
~~
Pietro Perona$^{1}$
\\\\
$^{1}$California Institute of Technology
\\
$^{2}$Center for Humans and Machines, Max Planck Institute for Human Development
\\
$^{*}$Equal contribution
}

\maketitle

\begin{abstract}
Measuring the accuracy of face recognition (FR) systems is essential for improving performance and ensuring responsible use. Accuracy is typically estimated using large annotated datasets, which are costly and difficult to obtain. We propose a novel method for 1:1 face verification that benchmarks FR systems quickly and without manual annotation, starting from approximate labels (e.g., from web search results). Unlike previous methods for training set label cleaning, ours leverages the embedding representation of the models being evaluated, achieving high accuracy in smaller-sized test datasets. Our approach reliably estimates FR accuracy and ranking, significantly reducing the time and cost of manual labeling. We also introduce the first public benchmark of five FR cloud services, revealing demographic biases, particularly lower accuracy for Asian women. Our rapid test method can democratize FR testing, promoting scrutiny and responsible use of the technology.
Our method is provided as a publicly accessible tool at \href{https://github.com/caltechvisionlab/frt-rapid-test}{https://github.com/caltechvisionlab/frt-rapid-test}.
\end{abstract}
\vspace{-2em}
\section{Introduction}
\label{sec:intro}

\begin{figure}[t!]
    \centering
    \includegraphics[width=\linewidth]{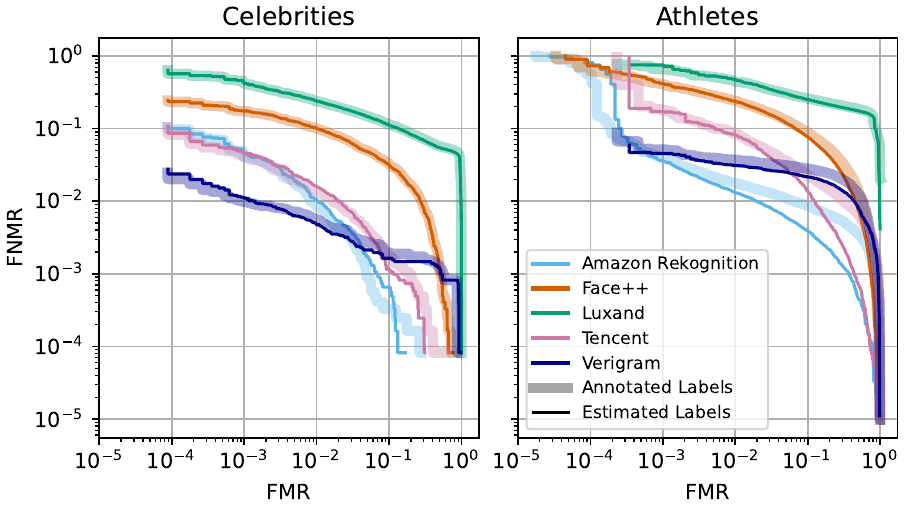}
    \caption{{\bf Unsupervised accuracy estimates on five FR cloud services match supervised estimates.} The plots show the False Non-Match Rate (FNMR) (equivalently, the False Reject Rate) vs. the False Match Rate (FMR)  (equivalently, the False Accept Rate) of five commercial cloud services on two collections of face images: Celebrities and Athletes. The thin dark lines indicate accuracy as estimated by our automated method. The thick pale lines indicate the ground-truth estimates through human labeling, which took more than a month of human labor to produce. See also~\cref{sec:accuracy-and-bias}.}
    \label{fig:accuracy}
\end{figure}

\begin{figure*}[t!]
    \centering    
    \includegraphics[width=.98\textwidth]{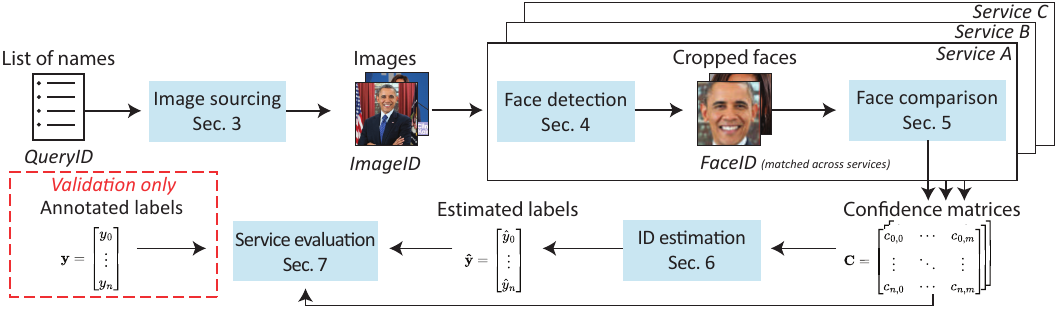}
    \caption{{\bf Overview of our method.} An operator provides a list of people's names that are used as queries for image URL sourcing from the internet. The images are accessed through their URL and are not stored. Several face recognition services are evaluated simultaneously (five in this study). Each service detects faces in the selected images and assigns a same-identity (or ``match'') confidence value to pairs of faces. From this information, an estimate of which faces belong to which identity is computed. Using this estimate FNMR-vs-FMR curves and bias estimates may be produced (\cref{fig:accuracy,fig:bias-equal-error}) to estimate the accuracy of each service. Our method does not require hand-annotation and estimates identity labels for each face image from the data.  In this study hand-annotated labels were collected purely to validate our method and were not available to our method.
    }
    \label{fig:overview}
    \vspace{-3mm}
\end{figure*}

Face recognition technology (FRT) is a convenient, no-contact, fast, accurate, and inexpensive way to interface securely people and machines.  From logging into our smart devices to boarding a plane, crossing a border, and finding missing children, FRT can make our lives more convenient and safer. Conversely, FRT misuse is possible and may lead to loss of privacy and violation of civil rights~\cite{van2020ethical,castelvecchi2020facial}. As applications increase, it is crucial to understand FRT's potential and implications. In particular, characterizing FRT systems' accuracy and bias is fundamental to informing developers, users, the public, and regulators about the merits and downsides of the technology and to improve it if necessary~\cite{chiroro1995investigation,hills2013eye,johnson2005we,kearns2019ethical,castelvecchi2020facial}. On the positive side: 
Accuracy in FRT systems has improved dramatically in the past five years. Today, systems achieve super-human accuracy~\cite{deng2019arcface,liu2017sphereface,schroff2015facenet,zhao2003face} and outperform even expert face analysts~\cite{phillips2018face}, with the potential to make our life more convenient and help reduce the harm that is currently caused by human error~\cite{thompson2000,albright2017eyewitnesses}. Furthermore, measuring and mitigating bias in algorithms is both feasible and effective, while measuring and correcting biases in human operators is notoriously difficult and can take a long time~\cite{chang2019mixed,mullainathan2019biased}. Additionally, applications of FRT to policing could work alongside DNA testing to help solve crimes quickly~\cite{harrisSpectrum21}, reduce bias in the justice system~\cite{meissner2001thirty}, and reduce the rate of wrongful identification~\cite{albright2017eyewitnesses,albright2021us} and imprisonment~\cite{thompson2000}. Thus, AI and FRT may become powerful agents of progress towards more fair, accountable, and transparent institutions~\cite{kleinberg2018discrimination,mullainathan2019biased}. Amongst the potential downsides: willful or inadvertent misuse, inaccuracy, and bias in face recognition technology could inflict harm on individuals and lead to social inequities~\cite{kearns2019ethical,van2020ethical,castelvecchi2020facial}. Responsible practice in developing and deploying the technology starts with measuring algorithmic accuracy and bias.

Unfortunately, testing FRT is expensive and laborious, and thus, it is not within the reach of most organizations. The best data on FRT accuracy is published by the U.S. National Institute of Standards and Technology (NIST). While the NIST team is reputable, experienced, and uses first-rate test sets (Sec.~\ref{sec:previous-work}), the situation is not ideal. First, not all FRT vendors submit their software to NIST for testing. Second, NIST is not testing cloud services. Third, to prevent vendors from overfitting, the test sets are kept secret and cannot be checked by independent experts for correlations and confounds. As a result, acceptance of NIST's figures rests on the institution's reputation rather than peer review.
A better state of affairs would be for any interested party to carry out those tests that it deems important independently. Today, this is not possible: the use of public datasets often leads to evaluation data leaking into the training process; new test sets collected by independent and academic teams are subject to privacy and copyright issues and are either too small or the ground truth is noisy at best, making testing results unreliable.

We propose a method to address these shortcomings and democratize the testing of FRT systems. It is based on two main ideas. First, use public data which is open to verification. To minimize the risk that the test data was used in model training, our method will only make use of images that have been made public recently. 
To minimize use risks images are analyzed on the fly and not stored.
Second, to make the testing practical and affordable and to avoid human labeling errors, our method does not rely on hand-labeled images but rather infers the ground truth labels of face identity from the algorithms' confidence values. This is a delicate algorithmic step that sits at our method's core.  
\cref{fig:overview} shows an overview of our proposed method.
Our main contributions are: (a) An inexpensive and practical observational method for accurately benchmarking face recognition algorithms, (b) a thorough experimental validation of our method, and (c) the first public benchmark of accuracy and bias for face recognition services in the cloud, with a side-by-side comparison of five popular cloud services.

\section{Previous work}
\label{sec:previous-work}

Estimating FRT accuracy and bias requires large, accurately labeled datasets to ensure tight confidence intervals. Furthermore, one needs diverse attributes representative of the general population to explore effects on all demographics.  A team with the U.S. National Institute of Standards and Technology (NIST)~\cite{grother2014face,grother2019face} has, over the past 20 years, developed state-of-the-art test datasets and testing practices. They test algorithms on six large ($\simsym 10$M images) datasets collected from visas, visa applications, border crossings, arrest mugshots, kiosk images, and images collected in the wild. Accurate identity annotations are achieved by combining trained government officials and identity documents. NIST publishes updated reports every few months on  NIST's  ``Face Recognition Vendor Test'' web page~\cite{nist_frvt}. A number of academic teams are also engaged in testing FRT~\cite{albiero2020analysis,albiero2020does,albiero2021gendered,krishnapriya2020issues}. They use public datasets that may have been included in the training sets of FRT vendors and whose identity labels are often not accurate. Thus, while valuable for science, academic tests may not be suitable for probing the accuracy of commercial systems.

Only governments and large tech companies have access to large, accurately labeled datasets. 
To democratize the testing of FRT algorithms, we need to reduce the cost of ground-truth identity annotation dramatically.
This has long been considered unlikely since accurate face identification using human annotators is very difficult~\cite{phillips2018face}, and benchmarking without an independently annotated ground truth might seem impossible. Semi-automated clean-up methods have been proposed to reduce the cost of improving label quality in training sets. The state-of-the-art, WebFace260M~\cite{webface260}, iteratively employs a face recognition (FR) model to clean the images and then re-trains that model with the new clean dataset to obtain an improved model. This dynamic has three shortcomings in our application: (i) Constructing a high-quality embedding requires millions of images, which is the case for training sets. We focus on test sets, which are a couple of orders of magnitudes smaller and yet require higher accuracy. (ii) The stable point of the method may suffer from partial mode collapse due to errors in the original data leaking into the model. This has been documented on small datasets~\cite{shumailov2024ai} and is difficult to verify on large datasets. (iii) Legislation in many countries prohibits the storage of biometric data that enables the identification of individuals. To address the first two problems, our method takes advantage of the ensemble of the embeddings of the systems being tested, which is high-quality and stable. To address the latter, our method does not store any identity labels or images, relying solely on the scores of FR systems.

Recent studies propose semi- and unsupervised methods for benchmarking algorithm accuracy~\cite{welinder2013lazy,ji2020can,chouldechova2022unsupervised}, which can estimate algorithms' accuracy even without access to an externally provided ground truth. These methods take advantage of statistical regularities of the confidence values of classification algorithms to estimate the underlying error statistics. We take inspiration from this work. We note that face recognition algorithms are highly accurate and thus will produce strongly bimodal distributions of confidence values when confronted with an (unlabeled) mix of same-ID and different-ID pairs of face images. We exploit this fact to estimate the ground truth image identity labels. 

Our method addresses the practical case of image collections obtained through web image searches, where face identities are typically correct in the 20-80\% range. Thus, our method is supervised in that identity (ID) labels are provided. However, crucially, it is designed to tolerate highly noisy face identity labels and does not require human supervision, such as hand-labeling, to correct such errors. In other words, we address the situation that lies in between the availability of exact ground truth (conventional benchmarks based on carefully annotated test sets) and the (quasi) complete absence of identity labels (previous unsupervised and semi-supervised methods~\cite{welinder2013lazy,chouldechova2022unsupervised}). 

 Like other benchmarks, our method uses {\em observational} test sets with demographic annotations (typically age, gender, ethnicity)~\cite{ricanek2006morph,liu2015deep,grother2019face,karkkainen2021fairface}. Recent literature indicates that observational methods are susceptible to bias from unmodeled confound variables and propose {\em experimental} approaches based on synthetic images \cite{balakrishnan2021towards,liang2023benchmarking}. We agree that there is merit in this concern. For the time being, we believe that both observational and experimental methods need to be used to assess the accuracy of FRT.

\section{Image sourcing} 
\label{sec:image-sourcing}

The procedure we recommend is designed to involve the minimum amount of human curation during image sourcing and does not create a static test image dataset. Instead, it selects images on the web, feeds pairs of them to the FRT cloud services being benchmarked, and retains only the resulting confidence values for analysis.

The process starts with a human-generated list of names that serve as queries in an image retrieval system.
For bias analysis, one can additionally provide demographic attributes (e.g. race, gender, age) for each name. For the evaluation part of this research, we generated two test sets with different image statistics. The first, {\em Celebrities}, starts from a list of names of famous people compiled by one of us. It contains 10 names in each of the eight demographic categories. 
The second, {\em Athletes}, is a subset of Wikipedia's list of 2020 Tokyo Olympic athletes. 
Our list comprises 2755 names and aims to be balanced across six demographics. Detailed statistics for both datasets can be found in \cref{appendix:image-sourcing}.

Only images published shortly before testing are considered to reduce the chance that test data was used to train the models being tested. In our experiments, URLs of the images were obtained from the Google Images API and the Google News API. 
Our script obtained a total of 5k images (an average of 67 per ID) for the Celebrities dataset and 223k images (an average of 81 per ID) for the Athletes dataset, leaving 2.2k and 58.6k, respectively, after face detection (see Sec.~\ref{sec:face-detection}).
The number of images found per identity varies significantly (see \cref{fig:hist-n-images-per-id}).
All images obtained using a given name string were assigned the same \textit{QueryID} (abbreviated as $q$ in the following). At this point, some, but not all, of the faces in the pictures obtained belong to the person whose identity corresponds with the search query. E.g., we expect that a search for ``Barack Obama'' will yield images of Barack Obama, as well as Michelle Obama, Joe Biden, and other world leaders. The algorithm described in Sec.~\ref{sec:gt_estimation} is designed to clean up these noisy labels.

To validate the results of this study, we manually added ground truth identity labels to each face image: one of the authors assigned label $y=1$ if the identity matched the query name, $y=0$ if not, and $y=-1$ for rare cases where the identity could not be confirmed even after meta information was consulted. The manual annotation process took a total of 200 hours, about 12 seconds per image on average.
Additional details about the image sourcing and annotation process can be found in \cref{appendix:image-sourcing}.

\section{Face detection}
\label{sec:face-detection}

Face detection, i.e. computing bounding boxes around each visible face, was carried out in every image using each one of the cloud services we tested. Images were sent to each cloud service's face detection API in our benchmark. Since bounding box sizes vary by service, we retained service-specific crops, assuming matching models were trained on these.

To establish a unified \textit{FaceID} across providers, we grouped bounding boxes of detected faces from different services. For this, we computed the IoU (Intersection over Union) metric for all pairwise combinations of detected faces in an image, excluding pairs of faces detected by the same provider. Pairs of detected faces with sufficiently high IoU measures (we used IoU $>0.2$) were grouped using Kruskal's algorithm~\cite{kruskal1956shortest} to compute the minimum spanning tree. Thresholding the minimum spanning produced face groups across services. Finally, each group was assigned a \textit{FaceID}.

We use the additional constraint only to include images where each provider found exactly one face. This allows us to include services that do not offer a face detection API and, therefore, can only include single-face images (Verigram in our case, see \cref{appendix:service-interrogation} for details on how to interrogate services without prior face detection). 
A positive side effect of using single-face images only is that lower-resolution background faces are less likely to occur, leading to more identifiable face crop images. 

\begin{figure*}[t!]
    \centering
    \includegraphics[width=\textwidth]{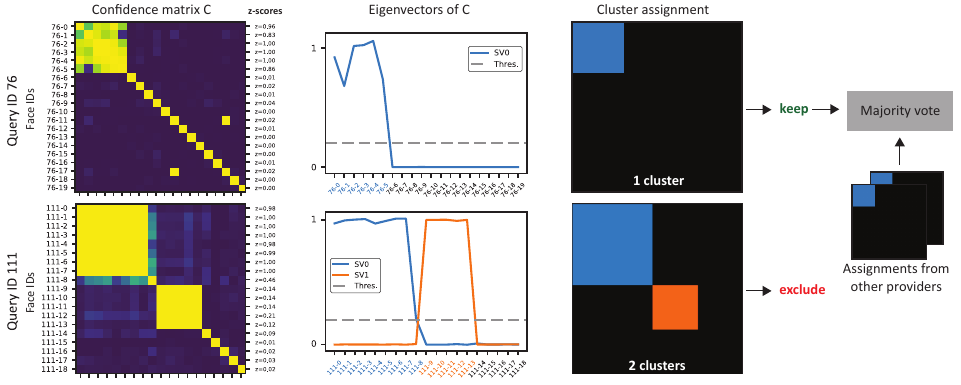}
    \caption{{\bf ID Label Estimation Method (\cref{sec:gt_estimation}).}  
    (Left column) Matrices showing the confidence values assigned by one of the FRT services to face pairs in queries $q=76, 111$. Each row and each column corresponds to a face image, and each matrix entry indicates the service's confidence that the corresponding pair of face images belongs to the same person (the indices have been rearranged to make the block structure apparent). Yellow indicates high confidence, and blue indicates low confidence. The top matrix has a single block, while the bottom one has two blocks, suggesting that two different identities with a significant number of images are associated with the query.
    (Second column) The top eigenvectors (singular vectors whose singular value exceeds a threshold) of the matrices, where the x-axis indicates the image index, act as indicator functions of which images are associated with each identity.
    (Third column) By thresholding the eigenvectors, the algorithm discovers which images belong to which identity. 
    (Right column) In the top row, information from the eigenvector is combined with corresponding eigenvectors from other services by majority vote. The bottom row does not meet the criteria for inclusion since it contains more than one identity and is discarded from further consideration.}
    \label{fig:gt-estimation}
    \vspace{-2mm}
\end{figure*}

\section{Face matching confidence scores}
\label{sec:face-comparison}

Face recognition cloud services assign a {\it confidence score} $C_{i,j}$ to each pair of faces $(i,j)$. A high confidence score indicates that the pair of faces are likely to belong to the same person, while a low confidence score indicates they are likely to belong to different people.  Estimating a service's accuracy requires computing the false non-match rate (FNMR, a.k.a. false reject rate) vs the false match rate (FMR, a.k.a. false accept rate) as a function of a minimum confidence threshold. Thus, for each cloud service provider and each face pair in the test set, we obtain pairwise confidence matches and evaluate the quality of such matches (Sec.~\ref{sec:experiments}) vis-a-vis the estimated labels (Sec.~\ref{sec:gt_estimation}).

FRT providers we tested are Amazon Rekognition~\cite{amazonRekognition}, Face++~\cite{face_plus_plus}, Luxand~\cite{luxand}, Tencent~\cite{tencent}, and Verigram~\cite{verigram}. We used paid services through regular subscriptions, except for Verigram, for which we received complimentary access for research purposes.
Computing confidence scores for all pairs of faces in the dataset is very expensive and unnecessary. Our method requires same-query pairs to evaluate FNMR as well as a comparable number of cross-query pairs to evaluate FMR. The set of face pairs to be evaluated by cloud providers was selected as follows:
1. All pairs of faces with the same $q$ were used both for face ID label estimation (\cref{sec:gt_estimation}) and model evaluation (\cref{sec:results}).
2. A random sample of pairs of faces with different $q$ and from the same demographic group was used for model evaluation (\cref{sec:results}). We sampled as many different-query pairs as same-query pairs.

A bimodal distribution of confidence values is expected from the services, where one mode is associated with different-ID (or impostor) matches, and one mode is associated with same-ID (or genuine) matches. Since the distribution of the confidence values, and thus the modes of the distributions, are different for each service, for estimating the identity labels (Sec.~\ref{sec:gt_estimation}), we normalized the confidence values to the $(0,1)$ range by mapping the range of values linearly so that the two modes are mapped to  0.0 and 1.0, clipping all smaller and larger values, respectively. Modes can be specified manually or estimated by fitting a bimodal Gaussian Mixture model to the confidence value distribution. For service evaluation (\cref{sec:results}), we report results using the services' original confidence values.

\section{Identity label estimation}
\label{sec:gt_estimation}

The next step is estimating the {\em identity label} $\hat{y}_i$ for each face image $i$. This requires two steps: deciding which identity (i.e., which physical person) corresponds to the query $q$ and deciding whether image $i$ corresponds to that person. Amongst the faces that were downloaded using the search string $q$, many will actually belong to different identities (Fig.~\ref{fig:correct-ID-scatter-plot}). Which person is the {\em correct identity} for a given name query $q$? Many people may be associated with the same name. How is this ambiguity resolved? For the hand-annotated labels, the ``correct identity'' is decided by the annotator. For our estimation method, the person/identity whose faces are prevalent in the set associated with $q$ is defined as the {\em correct identity}. The two criteria coincide almost always.

We describe an algorithm that estimates which identity is prevalent, i.e., it decides which is {\em correct identity} and estimates the corresponding face images. The end result is an estimate of the identity label for every image in each name. Such identity labels will be used to estimate the error rates for each service (Sec.~\ref{sec:results}). We start with an intuitive description of the algorithm's steps, and we give a more formal description of the algorithm at the end of the section.

The intuition for our algorithm is simple: pairs of faces in query $q$ corresponding to the same person will often, although not always, receive high pairwise confidence $C_{ij}^{qs}$ from service $s$. If the confidence is low, chances are that a third image $k$ of the same person will have high pairwise confidence with $i$ and $j$. Thus, we may use $C^{qs}$ as an {\em affinity} estimate to be used for grouping such faces using spectral factorization~\cite{shi1997normalized}. The largest group is most likely associated with the correct identity. Our algorithm is shown in Fig.~\ref{fig:gt-estimation}. For the sake of simplicity, consider first the most common case: the collection of images associated with a name consists of images that belong to the {\em correct identity} and other images corresponding to a sprinkling of different identities (Fig.~\ref{fig:gt-estimation}, first row). In this case, after reordering w.l.o.g. the image indices so that the correct identity is assigned contiguous indices, the confidence matrix $C$ is block-diagonal, with one large block with $C_{ij}\simeq 1$ where both $i$ and $j$ correspond to the correct identity, and the rest of the entries are $C_{ij}\simeq 0$ (Fig.~\ref{fig:gt-estimation}, left). Of course, the entries on the main diagonal are $C_{ii}=1$ since they correspond to the confidence of an image matching itself. 

The indices of the best estimate for the correct identity can thus be discovered automatically: it is well-known that the first eigenvector of such matrix is the vector $z$ where $z_i \simeq 1$ for $i$ corresponding to the correct identity and $z_i \simeq 0$ for $i$ corresponding to the other identities~\cite{shi1997normalized,perona1998factorization,shi2000normalized,zelnik2004self}.

Two more challenging situations are possible. The first is where more than one identity is present in the collection and associated with multiple images (Fig.~\ref{fig:gt-estimation}, 2nd row). In this case, multiple $z_i\simeq 1$ blocks are present in $C$. As discussed in the {\em spectral factorization} literature~\cite{perona1998factorization,shi2000normalized,zelnik2004self}, each block corresponds to a distinct eigenvector of $C$ that is associated with a large eigenvalue (the size of the eigenvalue is proportional to the size of each block) and may thus be identified automatically. The second insidious case occurs when no IDs are prevalent, i.e., when no identity is represented by at least a few images. In this case, no block-diagonal structure is detected, and all eigenvalues are small.

We may thus rely on two signals to automate the analysis of the confidence matrix $C$~\cite{perona1998factorization}. First, the magnitude of the eigenvalues of $C$ indicates the size of the corresponding blocks. The largest eigenvalue indicates the block that is associated with the {\em correct identity}. If the largest eigenvalue is small, that indicates that no identity is prevalent. Second, the entries of the eigenvectors of $C$ are non-negative for the eigenvectors corresponding to the unit blocks. Thus, eigenvectors with negative entries may be excluded from consideration. There is one rare exception: when two blocks have the same size (i.e., two distinct identities appear in equal numbers), the corresponding eigenvectors will be an arbitrary linear combination of the ideal non-negative eigenvectors and thus may contain negative entries. This case may be resolved automatically by rotating these {\em twin} (or triplets, or more) eigenvectors to obtain the canonical representation where all entries are non-negative.

We exclude test cases where a single identity cannot be reliably associated with a name. 
This includes the case where a single identity might not be represented by a single coherent cluster of face images. By discarding this challenging case, our algorithm is expected to slightly overestimate models' accuracies.

\textbf{Algorithm.}
In sum, our algorithm to estimate face identity labels relies on simple linear algebra operations on $C$. First, for each $q$ and $s$, the eigenvalue decomposition of the corresponding confidence matrix $C^{qs}$ is computed (since $C$ is symmetric, this is the same as computing the principal component analysis of the matrix). Second, the eigenvalues that exceed a threshold $T=4$ are selected. This guarantees that at least 4-5 faces are associated with the {\em correct identity}. If only one eigenvalue meets this criterion, call  $z^{qs}$ the corresponding eigenvector, provided that its entries are non-negative. If either multiple eigenvectors or none exceeds $T$ or some of the eigenvector entries are (non-trivially) negative, we discard the name $q$ and the corresponding image collection from further consideration.

At this point, we have computed vectors $z^{qs}$ for each remaining name $q$ and service $s$. Each such vector contains information on the face identity label, where $z_i \approx 1$ means that image $i$ belongs to the correct identity, while $z_i \approx 0$ means that image $i$ is a spurious identity. We now need to consolidate such estimates across services $s$.
First, we exclude those names $q$ where we do not find exactly one identity for all services $s$. Second, we exclude names where the number of faces associated with the prevalent identity is smaller than 5. Excluded faces are labeled $\hat{y}_i=-1$. Third, for all included faces, we determine the final estimated label $\hat{y}_i$ by majority voting across services, where we set a threshold $\tau = 0.2$ and, for a given $i$, if the majority of the $z_i^s > \tau$ over services $s$ then we assume that image $i$ is associated to the correct identity and set $\hat{y}_i=1$, and otherwise we set $\hat{y}_i=0$. We have empirically found that aligning the labels this way across services improves the alignment with the annotated face identities for all services (see \cref{fig:majority-vote} for details).

\vspace{0.75mm}
\textbf{Error types.}
Note that by excluding faces, we make our method's test set different from the test set that one would obtain from manual annotation. Thus, the accuracies obtained through the two methods may differ because of two distinct types of errors: \textbf{(A)} The error introduced by {\em excluding} certain queries from the estimation set and thus altering the underlying data distribution ($y_i \neq -1 \wedge \hat{y}_i=-1$); and \textbf{(B)} The error from misclassifying {\em included} faces ($y_i \neq \hat{y}_i \neq -1$). The impact of these two error types will be further studied in~\cref{sec:experiments}.

\section{Service evaluation}
\label{sec:results}

\begin{figure}[t!]
    \centering\includegraphics[width=.48\textwidth]{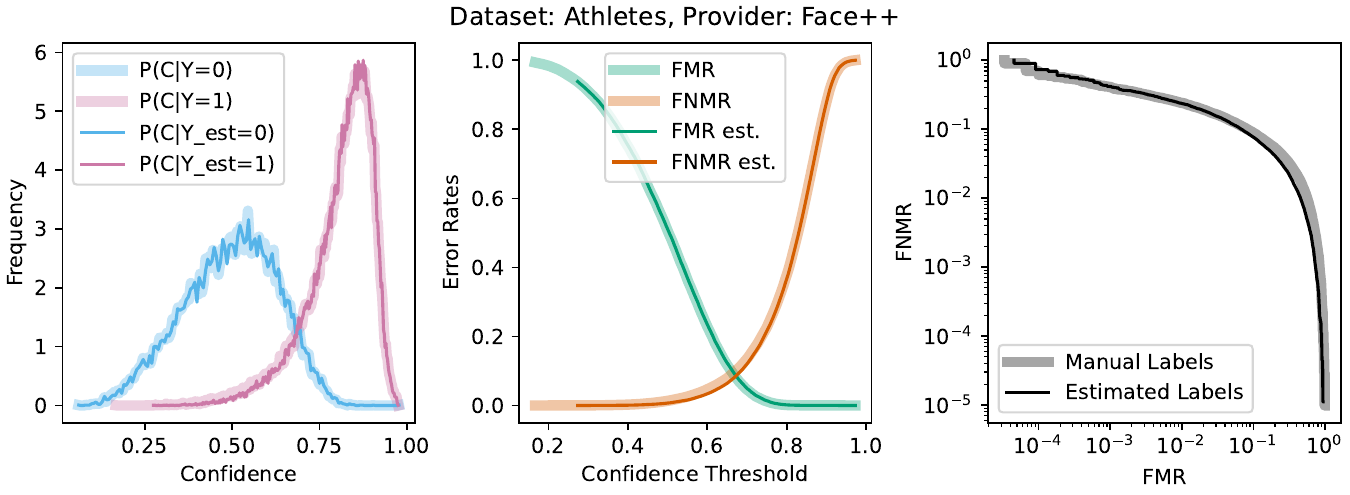}
    \caption{{\bf Sample of service output and evaluation.}  (Left) Distributions of confidence values for same-ID face pairs (pink) and different-ID pairs (blue) for the Face++ FRT service. (Mid) FMR and FNMR curves as a function of confidence thresholds. (Right) FMR-FNMR curves. Our method's estimate (thin dark lines) is close to the values obtained through hand-annotation (thick pale lines). These curves are incorporated in \cref{fig:accuracy} (left). Plots showing the same statistics separately for all datasets and all services may be found in Figs.~\ref{fig:results-all-celeb}, \ref{fig:results-all-athletes}.}
    \label{fig:fmr-fmnr-single-row-example}
    \vspace{2mm}
\end{figure}

To compute FNMR-vs-FMR curves, we need to divide each service's confidence values into two distinct sets: the ``genuine'' and ``impostor''  distributions. The genuine confidence values correspond to pairs of images that belong to the same identity. The impostor to pairs belonging to different identities. We do this twice: for our method's estimated identities and for the hand-annotated identities so that we may compare performance curves from our method with those from human annotation.  We only generate impostor pairs within the same demographic group since same-demographics impostors are the main challenge for FR services. Lastly, we demand that both cross-query images belong to the correct identity to guarantee that the two identities are different -- it is (remotely) possible that two images belonging to different queries but not to the correct identity actually belong to the same identity.  \cref{fig:fmr-fmnr-single-row-example} (left panel) shows these four distributions for one cloud service and one dataset.

To be clear,  we name $G$ and $I$ the genuine and impostor sets from hand-labeled identities, and $\hat{G}$  and $\hat{I}$ those from our method's estimates. We define the sets as follows, where $C_{ij}$ is the confidence value of an assessed face pair, $q$ is the QueryID, $d$ is the demographic group associated with $q$, $y$ is the annotated label, and $\hat{y}$ the estimated label:
\begin{align*}
    G &= \{ C_{ij} \mid (q_i = q_j) \,\,\wedge \,\, (y_i = 1) \,\,\wedge \,\, (y_j = 1) \}\\
    \hat{G} &= \{ C_{ij} \mid (q_i = q_j) \,\,\wedge \,\, (\hat{y}_i = 1)  \,\, \wedge \,\,  (\hat{y}_j = 1) \}\\
    I &= \{ C_{ij} \mid (q_i \neq q_j) \,\,\wedge\,\,  (y_i = 1) \,\,\wedge\,\,  (y_j = 1) \,\,\wedge \,\, (d_i = d_j) \}\\
    \hat{I} &= \{ C_{ij}\mid (q_i \neq q_j) \,\,\wedge \,\,  (\hat{y}_i =1) \,\,\wedge \,\, (\hat{y}_j = 1) \,\,\wedge \,\,  (d_i = d_j) \}
\end{align*}
(the notation is simplified to avoid clutter). Based on these distributions we derive the False Match Rates (FMR) and False Non-Match Rates (FNMR) as functions of the confidence threshold values (\cref{fig:fmr-fmnr-single-row-example}, mid panel) as well as FMR-FNMR curves (\cref{fig:fmr-fmnr-single-row-example}, right panel).

\vspace{0.75mm}
\textbf{Semi-supervised setting.}
While our approach is entirely unsupervised, in principle, it may be adapted to a semi-supervised setting by active sampling, i.e., by collecting human annotations just for a handful of the most ambiguous cases. These may be identified by looking at queries that were dropped by the estimation procedure and by picking samples with a high degree of ambiguity based on their z-scores (\cref{fig:gt-estimation}, left columns). One such analysis is shown in \cref{fig:semi-supervised}.

\section{Validation experiments}
\label{sec:experiments}

Does our method work? Does it correctly estimate the accuracy of face recognition systems? We validate our method by measuring the accuracy and bias of three face recognition services and comparing results to traditional hand-annotation. There are two main face recognition tasks: 1:1 matching and 1:n (one-to-many) matching. We focus on 1:1 matching, which is easier to analyze and thus the standard benchmark for accuracy~\cite{albiero2020analysis,phillips2018face-b,grother2019face}.

Two test datasets of face images, ``Celebrities'' and ``Athletes'', were defined, face-detected, and manually annotated following the steps described in Sec.~\ref{sec:image-sourcing} and Sec.~\ref{sec:face-detection}. 
The distribution of face image sizes is shown in Fig~\ref{fig:img-histogram}. The athletes' face images are, on average, smaller than the celebrities', and about 10\% are smaller than 64 pixels (harmonic mean of width and height), making recognition more challenging. The counts of \textit{correct identity} face images are shown in Fig.~\ref{fig:correct-ID-scatter-plot} (top row, x-axis) -- most of the queries yielded 10-40 {\em correct identity} useful face images for Celebrities, and 4-40 useful face images for Athletes. The fraction of \textit{correct identity} faces is mostly in the 60-90\% for Celebrities and 10-90\% for Athletes (Fig.~\ref{fig:correct-ID-scatter-plot}, bottom row, x-axis).

Five commercial cloud services (Amazon Rekognition, Face++, Luxand, Tencent, and Verigram) were used to compute confidence scores, as described in ~\cref{sec:face-comparison}. Histograms of the confidence scores we obtained are shown in ~\cref{fig:fmr-fmnr-single-row-example} (left) and Figs.~\ref{fig:results-all-celeb}, \ref{fig:results-all-athletes} (left). {\em While the plots show the confidence scores separately for the correct and false matches, our method has no access to this information and must estimate it.} 

Identity labels $\hat{y}$ were estimated following the algorithm presented in Sec~\ref{sec:gt_estimation}. Examples of label estimates for specific queries are shown in Fig.~\ref{fig:gt-estimation} and Fig.~\ref{fig:gt-estimation-supplementary}. Discrepancies between the number of positive labels $y_i=1$ in the hand annotations and estimated positive labels $\hat{y}_i=1$ are shown in Fig.~\ref{fig:correct-ID-scatter-plot}. The estimated labels mostly agree with the human-annotated labels ($\simsym 99.5\%$ agreement for Celebrities and $\simsym 97.8\%$ for Athletes, see also \cref{tab:confusion_matrix}). 
Disagreements may cause our method's estimate to differ from the estimate obtained from the hand-labeled faces (see below).

FNMR-vs-FMR curves are the end result of our process, as shown in Fig.~\ref{fig:fmr-fmnr-single-row-example} (left) (for Face++), and~\cref{fig:accuracy} (comparing the five services). These plots show the trade-off between false rejects and false accepts for each service. We compare curves obtained through hand-labeling of the dataset and through the estimation that is produced by our method. The fact that hand-labeling and our method produce similar curves shows that our fully automated method may be used to replace the manual (and very expensive) annotations. The discrepancies are caused by two factors: first, some of the queried names $q$ were discarded by the estimation method (\cref{sec:gt_estimation}, \textit{Type A error}), i.e., the sets of faces on which the services are tested are (slightly) different.  Second, as discussed above, there are a few disagreements $\hat{y_i} \neq y_i$ in the label estimates (\textit{Type B error}). Further evaluation shows that while the former is the main cause of discrepancy for the Celebrities dataset, the latter predominantly causes the error when evaluating the Athletes dataset (\cref{fig:achievable-acc}). Despite these differences, our estimates closely match hand annotations, slightly underestimating error rates.

\section{Accuracy and bias}
\label{sec:accuracy-and-bias}

\begin{figure*}[t!]
    \includegraphics[trim=0mm 1mm 0mm 0mm, clip,width=\linewidth]{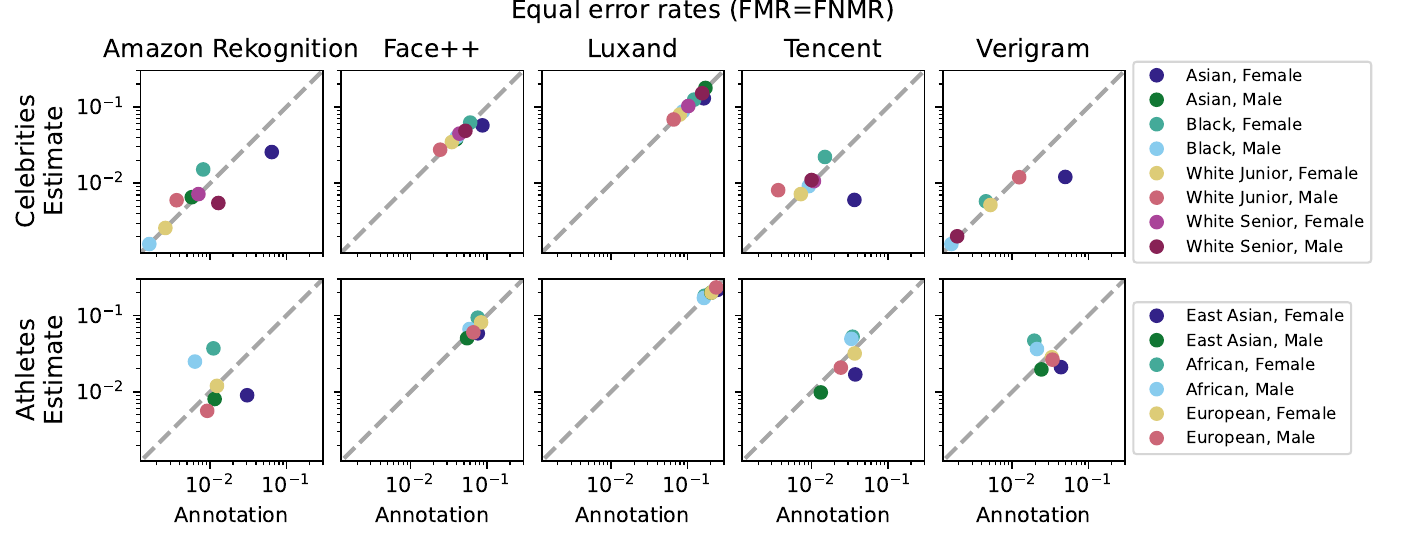}
    \caption{\textbf{Measuring bias vis-a-vis gender and race or geographical area}. Our method's (Estimate) vs hand-annotated (Annotation) equal error rate (FMR=FNMR) of the five services computed for each intersectional group defined by gender and race (Celebrities) or geographical area of country (Athletes). Our method correctly detects large biases (i.e., differences in accuracy across demographic groups): see the markedly higher error rates for Asian female celebrities in Amazon Recognition and Verigram, the two more accurate services. Detailed FMR-FNMR curves per demographic group are depicted in \cref{fig:bias}. Confidence intervals, computed using Wilson's method~\cite{fogliato2024confidence}, are about 3x larger than the markers (not shown to preserve visual clarity).}
    \label{fig:bias-equal-error}
    \vspace{-3mm}
\end{figure*}

The accuracy of the five services may be assessed from the FNMR-vs-FMR plots of \cref{fig:accuracy} as well as Figs.~\ref{fig:results-all-celeb}, \ref{fig:results-all-athletes}. The same conclusions on absolute and relative accuracy may be reached both from the hand-annotated test sets and from our method.  First, Luxand and Face++ services are markedly less accurate than the other three. Verigram is the most accurate on Celebrities at relevant FMRs (low FMR), and is a tad less accurate than Amazon Rekognition on Athletes. Second, all services are more accurate on Celebrities than on Athletes---this is expected since Celebrities have many well-lit posed photographs and overall good resolution, while the Athletes dataset contains challenging photographs taken during athletic events, where the subjects are wearing sports equipment such as goggles, are grimacing, and the poses are more challenging.

Each identity in our datasets was annotated for gender and race. Therefore, we can estimate demographic biases in the services we test. The FNMR-vs-FMR curves are shown disaggregated by demographic groups in Fig.~\ref{fig:bias}. To make it easier to understand the biases, we show the equal error rate (FNMR=FMR) for each curve in Fig.~\ref{fig:bias-equal-error} where each point corresponds to a demographic group, and the equal error rate estimated by our method is compared to the equal error rate that is computed using hand-labeling of the identities. It is clear from these plots that our method is able to estimate bias accurately when errors are large. When algorithmic errors and biases are small, estimate errors are proportionally larger, possibly due to smaller sample sizes.

Observational methods cannot resolve whether biases are in the algorithm or in the test data~\cite{balakrishnan2021towards} (see also in~\cref{sec:previous-work}). Since different bias patterns are revealed for Athletes and Celebrities (see~\cref{fig:bias-equal-error}, \cref{fig:bias}), it is prudent to assume that biases in the test data are prevalent here. 

\section{Discussion and conclusions}
\label{sec:conclusions}
\label{sec:discussion}

We have presented a novel method to estimate the accuracy and bias of face recognition services. Our method eliminates the need for hand-annotating the identity of faces in a test set, which is slow, extremely expensive, and can be inaccurate. Dataset annotation is the main blocker for anyone wishing to test face recognition systems' accuracy and bias. An attractive feature of our method is speed since each step is entirely automated after an initial source of names has been chosen.  A test, including forming a test set, obtaining confidence ratings from the services to be tested, and analyzing the data to estimate performance, will be completed in about one day ($\simsym2$k photos) to four weeks ($\simsym60$k photos). We estimate that the alternative, which includes collecting images not used to train face recognition models, as well as hand-labeling and hand-curation of the test set, may take many months. Thus, our method democratizes access to testing face recognition systems, a crucial activity in responsible AI.
Our method tests services at a certain point in time. Thus, one can discover when and whether a service has changed (see \cref{fig:celeb-aws-old-v-new} for such an analysis).
Our system uses face images as transient data and does not require persistent storage of images.
In addition, the method can easily be used by a trained operator to select and hand-annotate a fraction of images where the identity label is ambiguous, thus maximizing accuracy while minimizing the additional cost for annotation.

Using our method, we could estimate the error rates and biases of five cloud-based face recognition services quickly and accurately. To the best of our knowledge, this is the first published assessment of the accuracy of cloud-based commercial face recognition systems. The only other available measurements of commercial systems come from the National Institute of Standards and Technology (NIST), which does not directly test cloud-based services but rather relies on standalone code that is submitted to NIST by the vendors. 

We validated our method by comparing its estimates with those provided by hand-annotation and found a very close agreement for the Celebrities dataset and good-enough agreement for the more challenging Athletes dataset, where good-enough means that the same conclusions on absolute accuracy, relative accuracy, and bias may be reached.

Some steps in our method could be further refined. First, queries that yield multiple identities may be used rather than discarded since our label estimation scheme can handle multiple identities. Second, simultaneous testing of more than five services ought to improve the majority vote we use to estimate identity labels and thus further improve accuracy. 

Our method has limitations. 
First, in some contexts, using images published online may be undesirable. Second, identity estimation relies on the fact that face recognition services are accurate, and thus, the confidence values they produce are strongly bimodal. If this were not true, then our method would not work. Thus, our method is unsuitable for testing face recognition services in extreme conditions where face recognition models will struggle, e.g., images with very low pixel resolution, resolving the identity of identical twins, faces wearing surgical masks, and extreme age differences. In these conditions, it will always be best to resort to hand-annotated datasets.
Third, while our method is designed to reduce potential image overlaps between training and testing data, it does not address the issue of potential identity overlaps~\cite{wu2024identity}. 

\vspace{-3mm}
\vfill
\paragraph{Ethical considerations.}
Our goal is to enable greater transparency and public awareness of the accuracy of cloud-based face recognition systems (risks and benefits of face recognition technology are discussed in \cref{sec:intro}). The images processed by our method are used to evaluate the accuracy and bias of face recognition systems, not to train face recognition models. 
Our method is designed with privacy in mind \cite{lala2023data}, and no face image datasets are created or stored. Instead, images are processed on the fly, and any identifiable data is used only for image search and discarded afterward. Our tool will produce aggregated accuracy and bias metrics, minimizing privacy risks.

\vspace{-3mm}
\vfill
\paragraph{Acknowledgments.}
Ignacio Serna's research was supported by the UAM-Santander Young Researcher Program 2022.
We thank Verigram for providing access to their face recognition API for the purpose of this research.

\newpage
{
\small
\bibliographystyle{ieee_fullname}
\bibliography{refs}
}

\clearpage
\begin{appendices}
\onecolumn
\setcounter{figure}{0}    
\setcounter{table}{0} 
\setcounter{section}{0}
\renewcommand{\thesection}{\Alph{section}}%
\renewcommand\thefigure{S.\arabic{figure}} 
\renewcommand\thetable{S.\arabic{table}}
\begin{center}
    \begin{minipage}{\textwidth}
        \centering
        \fontsize{16}{20}\selectfont\bf
        \ifarxiv
        \else
        A Rapid Test for Accuracy and Bias of Face Recognition Technology\\
        \fi
        Supplementary Materials
        \vspace{12pt}
    \end{minipage}
\end{center}

\section{Image sourcing}
\label{appendix:image-sourcing}

\paragraph{Focus on recent images.} Our method tests cloud services on recent photos -- older photos are more likely to have been used for training by cloud providers, which would bias the results. Our method uses images that had been published on the web within 12 months before the test is run. 

Our method does not filter photos using EXIF data, which is typically present in the photograph's file and often contains the date on which the photo was taken. That is because we found EXIF data to be sometimes misleading and reports the date a photo was uploaded/hosted rather than the date it was taken. Additionally, some images do not carry EXIF data. 

\paragraph{Google News.} The process starts by inputting each name into the Google News Search API, which, in turn, yields up to 100 recent news articles related to the specified query. For each news article, the \href{https://github.com/codelucas/newspaper}{``newspaper" Python package} outputs a link for ``the best image to represent this article (the first image in the HTML markdown where the main article lies)." This process yields an average dataset of between 30 and 80 images per input name, depending on the popularity of the name. The variability in the dataset size is contingent upon the popularity of the individual's name and the corresponding availability of relevant images in the news articles. Overall, we found that few articles are available for individuals of Asian origin, and thus, this method for sourcing images may not work well in general.

\paragraph{Google Images.}  The image retrieval process involves issuing requests to the \href{https://developers.google.com/custom-search/v1/overview}{Google Custom SearchAPI} for each input name. Each query is designed to return a maximum of 10 items. The parameters of the request offer flexibility in specifying the number of results, their position, the date of the results, the result type (in this case, images), and more. To accumulate a total of 100 image links for a single name, a series of 10 requests is made, systematically varying the position of the results in each subsequent query. The variability of the number of images is contingent upon the popularity of the individual's name and the corresponding availability of relevant images.

\begin{figure}[ht!]
    \centering
    \includegraphics[width=.75\textwidth]{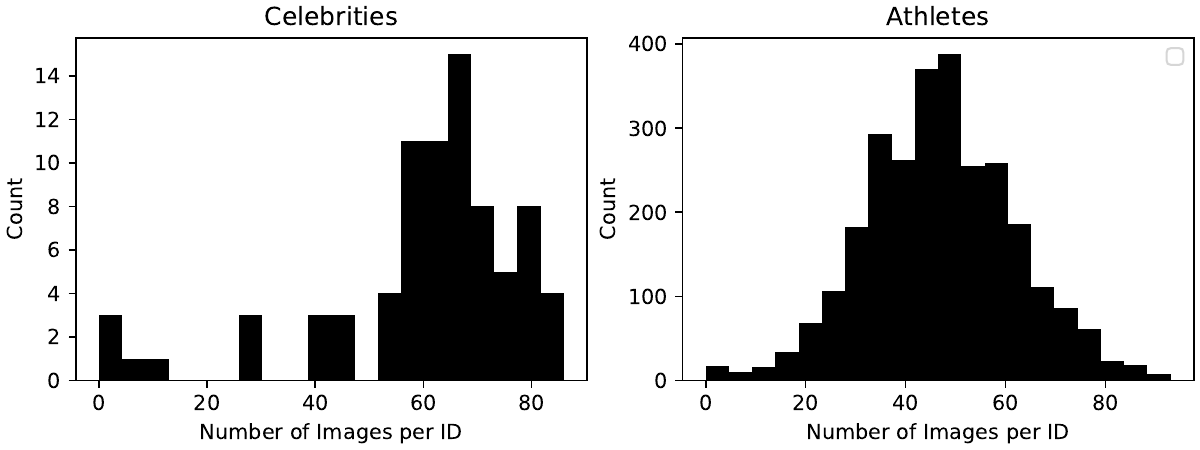}
    \caption{{\bf Number of images per identity.} The histograms show that, on average, more images were obtained for the Celebrities than for the Athlete datasets.}
    \label{fig:hist-n-images-per-id}
\centering
\subfloat[\textbf{Celebrities} N. of images]{\label{fig:celeb-downloaded}\includegraphics[width=0.5\columnwidth]{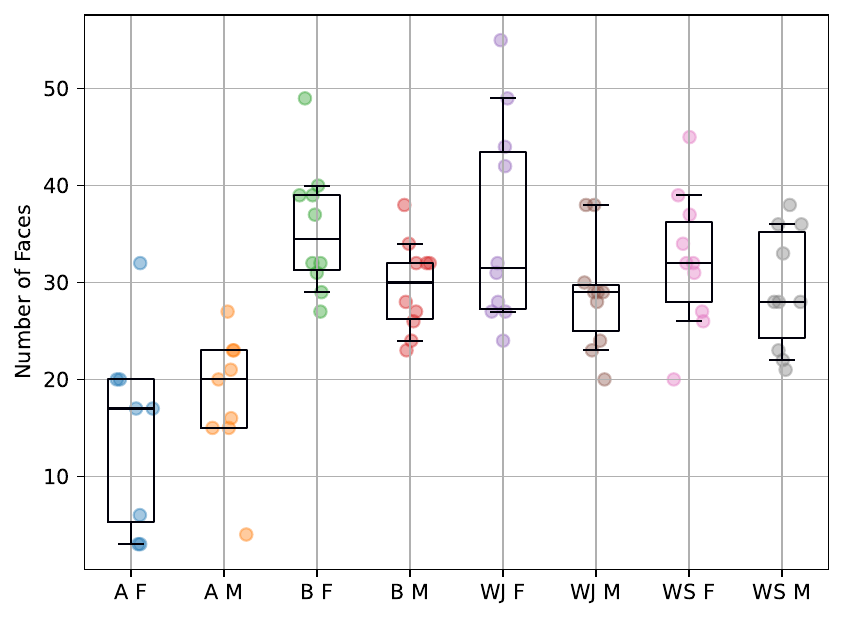}}\hfil
\subfloat[\textbf{Celebrities} \% of correct images]{\label{fig:celeb-correct}\includegraphics[width=0.5\columnwidth]{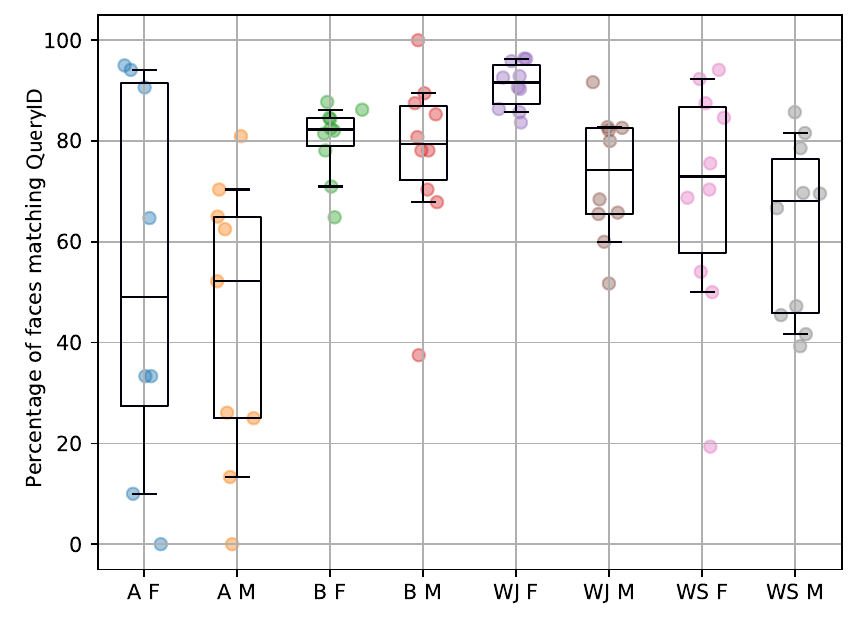}}\hfil
\caption[BiasAnalysis]{\textbf{Statistics for the Celebrities dataset.} (a) Number of face images per demographic group (nomenclature below). (b) Percentage of correct face images (the identity of the person in the image matches the name queried) per name by group, as established by hand labeling. Each marker represents an individual. The box plot spans the interquartile range (75th and 25th percentiles of the data), and the whiskers extend to the 90th and 10th percentiles. Nomenclature: F: Female, M: Male, A: Asian, B: Black, WJ: White Junior, WS: White Senior.}
\label{fig:celeb-dataset-statistics}
\centering
\subfloat[\textbf{Olympics} N. of images]{\label{fig:olympcis-downloaded}\includegraphics[width=0.5\columnwidth]{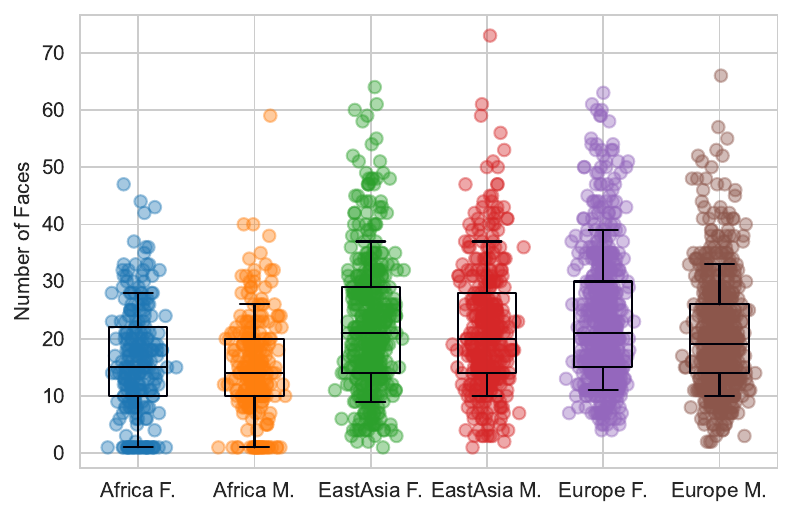}}\hfil
\subfloat[\textbf{Olympics} \% of correct images]{\label{fig:olympics-correct}\includegraphics[width=0.5\columnwidth]{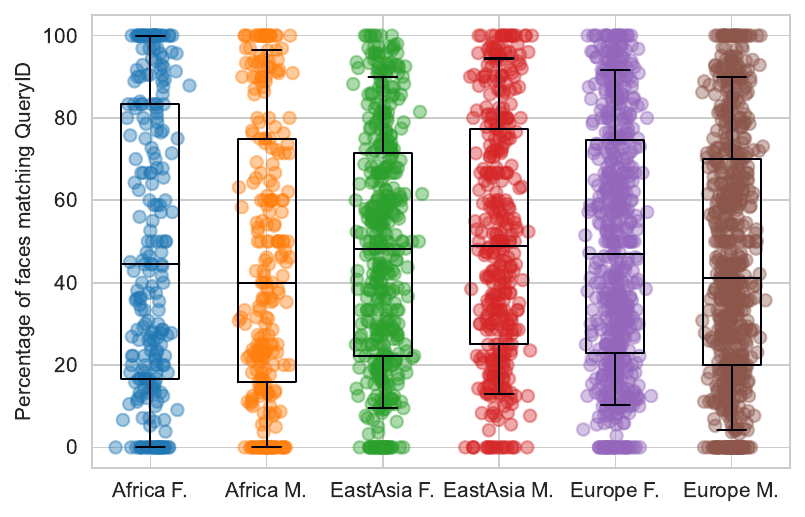}}\hfil
\caption[BiasAnalysis]{\textbf{Statistics for the Athletes dataset.} (a) Number of downloaded face images per name by group. (b) Percentage of correct face images (the identity of the person in the image matches the name queried) per name by group. Each marker represents one individual. The box plot spans the interquartile range (75th and 25th percentiles of the data), and the lines extend to the 90th and 10th percentiles. F. stands for Female, and M. for Male.}
\label{fig:athletes-dataset-statistics}
\end{figure}

\paragraph{Lists of names.} 
The full list of names and meta information for the two datasets sourced in this study can be found \href{https://github.com/caltechvisionlab/frt-rapid-test/tree/main/paper-supplement}{here}.
We experimented with two methods of generating the lists. One of the authors manually generated the list of celebrities (mostly singers and actors), balancing different demographics. The list of athletes was automatically generated by sampling from the  \href{https://en.wikipedia.org/wiki/Category:Nations_at_the_2020_Summer_Olympics}{2020 Summer Olympics Wikipedia page}.
The \textit{Celebrities} dataset was constructed using Google News with the individual's name as a search query. The \textit{celebrities} dataset includes 80 names and is divided into eight demographic groups. We compiled the list by selecting 10 names for each group, determined by gender (male/female), racial background (Asian/Black/White), and age (junior/senior) for Whites only. Demographic information on gender, age and race (for Celebrities) was obtained from Wikipedia and matched other public information. The number of images obtained for each identity is histogrammed in \cref{fig:hist-n-images-per-id} (left).
For the \textit{Athletes} dataset, we used Google Images for the above-mentioned reasons. The query was constructed using the athlete's name and country (i.e., ``<FirstName> <LastName> <Country>''). We did not use race information but rather the athletes' country's continent. The dataset contains 2755 names originating from 74 countries, strategically selected to achieve gender balance within three distinct ethnic origins: Africa, East Asia, and Europe.  The criteria for country selection included a deliberate effort to maintain an approximate equilibrium between males and females within three distinct ethnic origins: Africa, East Asia, and Europe. Notably, the chosen countries were characterized by historical homogeneity, ensuring a focus on regions where demographic mixing has been limited. The countries within each are: 

\begin{itemize}
    \item \textbf{Africa}: Angola, Bahamas, Bahrain, Barbados, Benin, Botswana, Burkina Faso, Burundi, Cameroon, Cayman Islands, Central African Republic, Chad, Democratic Republic of the Congo, Eritrea, Eswatini, Ethiopia, Gabon, Ghana, Grenada, Guinea, Guinea-Bissau, Guyana, Haiti, Ivory Coast, Jamaica, Kenya, Lesotho, Liberia, Madagascar, Malawi, Mali, Mauritania, Mozambique, Namibia, Niger, Nigeria, Republic of the Congo, Rwanda, Senegal, Sierra Leone, Solomon Islands, Somalia, South Sudan, Sudan, Tanzania, The Gambia, Togo, Trinidad and Tobago, Uganda, Zambia, Zimbabwe.
    \item \textbf{East Asia}: China, Chinese Taipei, Hong Kong, Japan, Mongolia, South Korea.
    \item \textbf{Europe}: Austria, Belarus, Belgium, Czech Republic, Denmark, Estonia, Finland, Germany, Iceland, Latvia, Liechtenstein, Lithuania, Norway,  Slovakia, Slovenia, Sweden, Ukraine.
\end{itemize}

The number of images obtained for each identity is histogrammed in \cref{fig:hist-n-images-per-id} (right).

\paragraph{Duplicate removal.}
Duplicate and quasi-duplicate photos are unnecessary and artificially distort accuracy estimates. We identify and remove duplicate images by computing their cosine similarity in MobileNet~\cite{howard2017mobilenets} embeddings, pre-trained on ImageNet. Pictures were organized into similarity groups if the cosine similarity was greater than 0.9, and the medoid image of each group was retained.
We eliminated duplicates twice: first, on sourced images and then again later for the cropped faces (Sec.~\ref{sec:face-detection}).

\paragraph{Challenges in identity consistency.}
A potential issue occurs when the same individual is inadvertently listed multiple times with variations in their name (e.g., “Barack Obama” and “Barack Hussein Obama”). This situation will result in overlapping sets of images treated as different identities for what is technically the same identity. Such overlaps will cause the evaluation process to underestimate model accuracy. It is the responsibility of users to avoid such duplicates.

\paragraph{Manual label annotation.}
Manual annotation was done using a self-developed browser-based interface.
All faces from each query were presented together one query at a time on the browser, with options to display either the full or cropped image (refer to Sec.~\ref{sec:face-detection}). 
A first pass was made based on facial appearance. Ambiguous faces, where it was otherwise impossible to identify the person solely based on the image, were reviewed, and a determination was made using meta information (from the website where the image was published or from image captions). There are rare cases of non-photorealistic images (e.g., signal processing filter, paintings, caricatures, pixel art) being assigned a positive label as well, as we have observed that FRT services usually manage to identify those correctly.  The manual annotation process took a total of 200 hours, about 12 seconds per image on average. The annotator responsible for labeling all the images is one of the co-authors and resides in Europe, potentially increasing the likelihood of annotation errors for Asian faces. To assess the accuracy of our labels, all disagreements between the manual annotation and our method's automatic ID assignment (Sec.~\ref{sec:gt_estimation}) were reviewed by two people, reducing the likelihood of errors to a very low level.  We specifically reviewed about 600 disagreements, revising approximately 500 for Athletes and about 10 for Celebrities (out of about 25,000 total labeled by our method). Thus, we estimate that our labels are $>98$\% correct for Athletes and $>99.5$\% correct for Celebrities.

\FloatBarrier

\section{Face detection}
\label{appendix:service-interrogation}

\begin{figure}[h!]
    \centering
    \includegraphics[width=.75\textwidth]{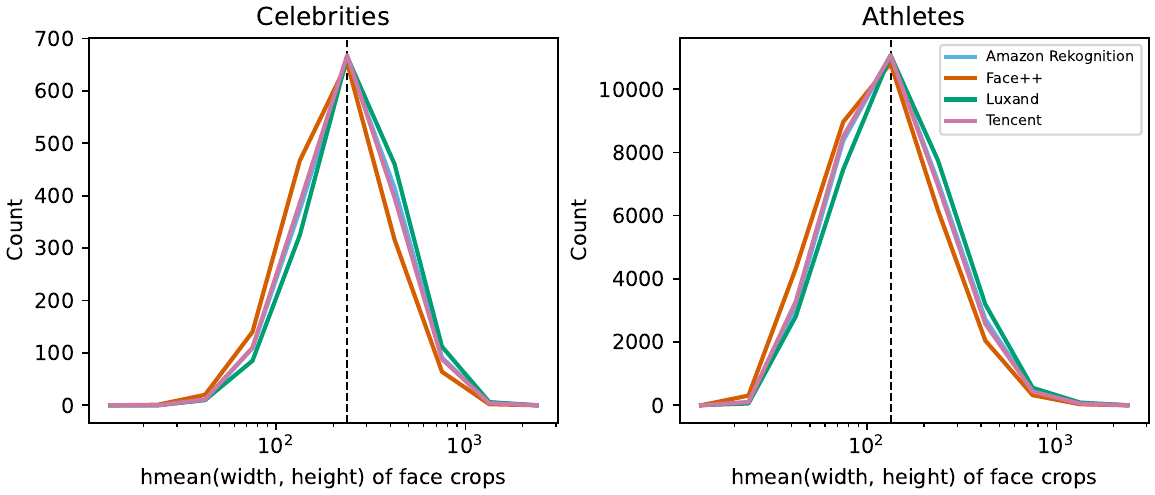}
    \caption{{\bf Histogram of face sizes.} The harmonic mean of width and height (in pixels) for each face image is used as a proxy of image size.  Only images that are included in the validation experiments (\cref{sec:experiments}) are considered here. Notice that the Athletes dataset face images have slightly lower resolution than the Celebrities dataset's. Face++ crops faces more tightly and Luxand more loosely than the other services.}
    \label{fig:img-histogram}
\end{figure}

\cref{fig:img-histogram} shows a distribution of the sizes of the face-bounding boxes for each provider. Not all services detect every face in an image. We only keep those face images that are detected by all services. Our method maps one bounding box per provider to a common FaceID as described in \cref{sec:face-detection}. Across all retrieved images, our method is able to assign $78.3\%$ (Celebrities) and $39.4\%$ (Athletes) of the detected bounding boxes faces to a common FaceID. 
If we additionally employ the restriction to only keep images that show exactly one face (see \cref{sec:face-detection}) we keep $19.4\%$ (Celebrities) and $39.0\%$ (Athletes) of the detected bounding boxes, respectively.

Some 1:1 face matching services might not offer a face detection service (e.g., Verigram). In this case, one can use the ``Multiple faces detected'' error response of the matching service to efficiently determine the subset of single-face images as follows:

\begin{algorithm}
\caption{Procedure to find images containing exactly one face for services without a face detection API.}
\begin{algorithmic}[1]
\State \textbf{let} valid\_imgs be a set
\State \textbf{let} invalid\_imgs be a set
\For{\textbf{each} (i1, i2) \textbf{in} image\_pairs}
    \If{i1 \textbf{in} invalid\_imgs \textbf{or} i2 \textbf{in} invalid\_imgs}
        \State \textbf{continue}
    \EndIf
    \State result = compare\_imgs(i1, i2)
    \If{result == "invalid"}
        \If{i1 \textbf{in} valid\_imgs}
            \State invalid\_imgs.add(i2)
        \EndIf
        \If{i2 \textbf{in} valid\_imgs}
            \State invalid\_imgs.add(i1)
        \EndIf
    \Else
    \State
    store\_confidence\_value(i1, i2, result)
        \State valid\_imgs.add(i1)
        \State valid\_imgs.add(i2)
    \EndIf
\EndFor
\end{algorithmic}
\end{algorithm}

\FloatBarrier
\section{Identity label estimation}
\label{sec:identity-label-estimation-supplementary}

\begin{figure}[ht!]
    \centering
    \includegraphics[height=.84\textheight]{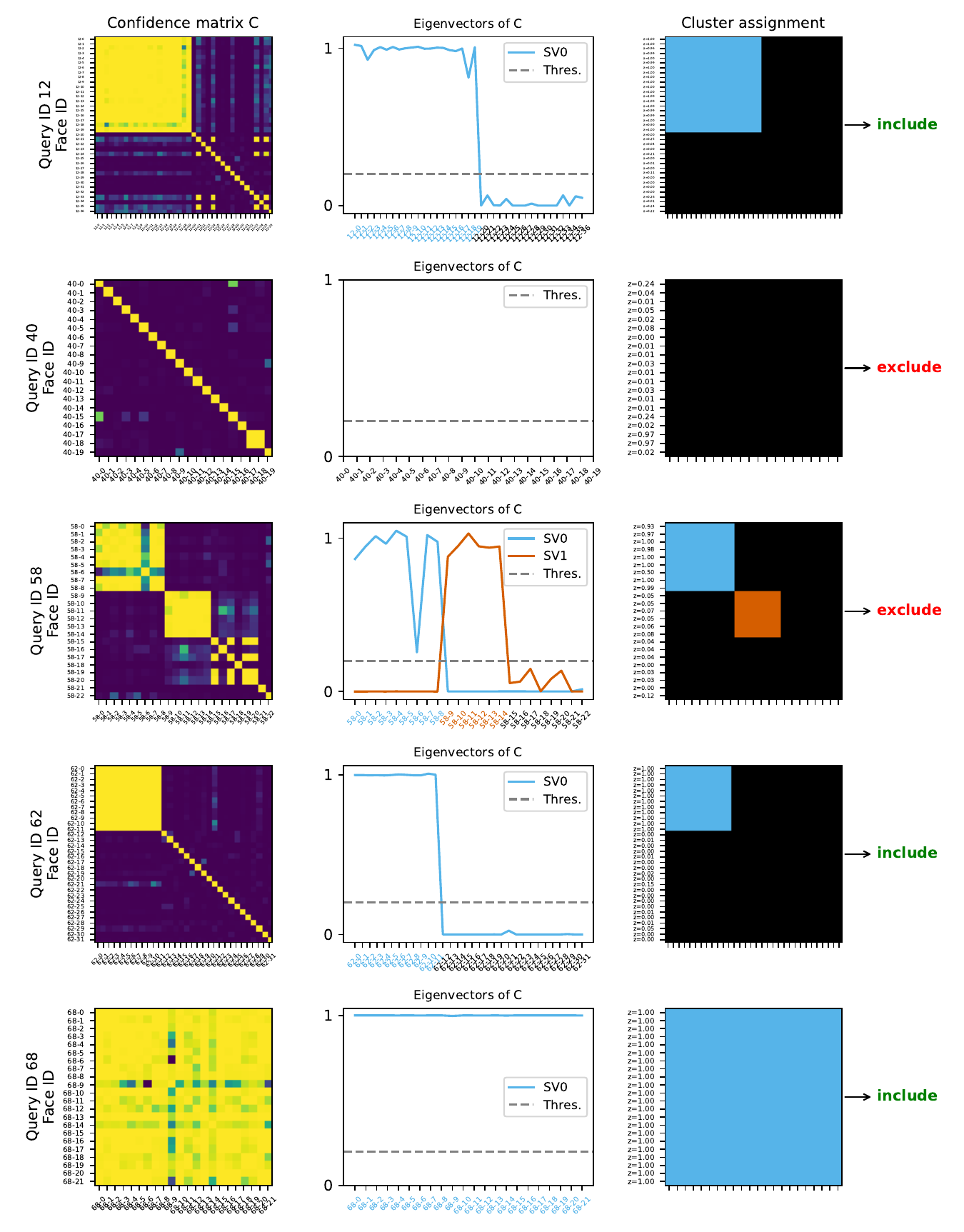}
    \caption{{\bf ID label estimation examples.} The procedure is described in~\cref{sec:gt_estimation} and this figure provides additional examples to supplement~\cref{fig:gt-estimation}. The first column shows the pairwise confidence matrix $C$  that was obtained from one service by comparing all pairs of faces that were associated with a given name query (reported on the left). The second column shows the eigenvector(s) of $C$ that meet the criteria described in~\cref{sec:gt_estimation}. The third column shows the groups, as well as the final decision, that are computed by our algorithm. The first row shows an easy case with about half the faces belonging to a dominant identity and the rest belonging to unrelated identities. The second row shows a case where all the identities are unrelated. The third row shows two, perhaps three, dominant identities (our algorithm recovers two). The fourth row is similar to the first row, with fewer images belonging to the dominant identity. The last row shows an easy case, where all the images are associated with the same identity.}
    \label{fig:gt-estimation-supplementary}
\end{figure}

\begin{table}[ht!]
\centering
\caption{{\bf Confusion matrices for annotations vs estimations.} See \cref{sec:gt_estimation} for details on how these labels are assigned. $y$ denotes the label that was assigned by hand, and $\hat{y}$ is the label that was assigned by our algorithm. $y=-1$ was assigned when faces were not unambiguously identifiable by the human annotator (e.g., occluded faces). In addition, we report the number of faces that were crawled but excluded from the analysis (``n excluded'') as they did not meet the minimum requirements: at least 8 crawled faces must be present per query, and all services must have been able to make an estimate based on the image.}
\label{tab:confusion_matrix}
\begin{subtable}{.33\linewidth}
\centering
\caption{Celebrities}
\begin{tabular}{l|ccc}
& $\hat{y}=1$ &$\hat{y}=0$ &$\hat{y}=-1$ \\ \hline
$y=1$ & 1213 & 2 & 422 \\
$y=0$ & 3 & 338 & 218 \\
$y=-1$ & 0 & 0 & 0 \\
\multicolumn{4}{l}{n excluded: 13} \\
\end{tabular}
\end{subtable}%
\begin{subtable}{.33\linewidth}
\centering
\caption{Athletes}
\begin{tabular}{l|ccc}
& $\hat{y}=1$ & $\hat{y}=0$ & $\hat{y}=-1$ \\ \hline
$y=1$ & 12311 & 117 & 16576\\
$y=0$ & 254 & 4351 & 21076 \\
$y=-1$ & 1 & 0 & 26 \\
\multicolumn{4}{l}{n excluded: 3913} \\
\end{tabular}
\end{subtable}
\begin{subtable}{.33\linewidth}
\centering
\caption{Celebrities 2024 (Appx.~\ref{sec:over-time})}
\begin{tabular}{l|ccc}
& $\hat{y}=1$ & $\hat{y}=0$ & $\hat{y}=-1$ \\ \hline
$y=1$ & 874 & 3 & 65\\
$y=0$ & 9 & 145 & 81 \\
$y=-1$ & 0 & 0 & 0 \\
\multicolumn{4}{l}{n excluded: 50} \\
\end{tabular}
\end{subtable}
\end{table}

\paragraph{Split-IDs.}
A practical challenge when using clustering-based methods for pseudo-annotation are clusters of images belonging to the same identity but marked as separate IDs. In practice, this can occur for individuals who undergo significant physical changes (e.g., due to facial surgery) or actors whose images are strongly associated with particular roles (e.g., an actor widely known as “Batman”). Our methodology addresses such cases by discarding identities that fail to form a single coherent cluster during the grouping process (see \cref{sec:gt_estimation} and \cref{fig:gt-estimation-supplementary}). By discarding these challenging cases our algorithm slightly overestimates the accuracy of the models.

\begin{figure}[ht!]
    \centering
    \includegraphics[width=0.49\textwidth]{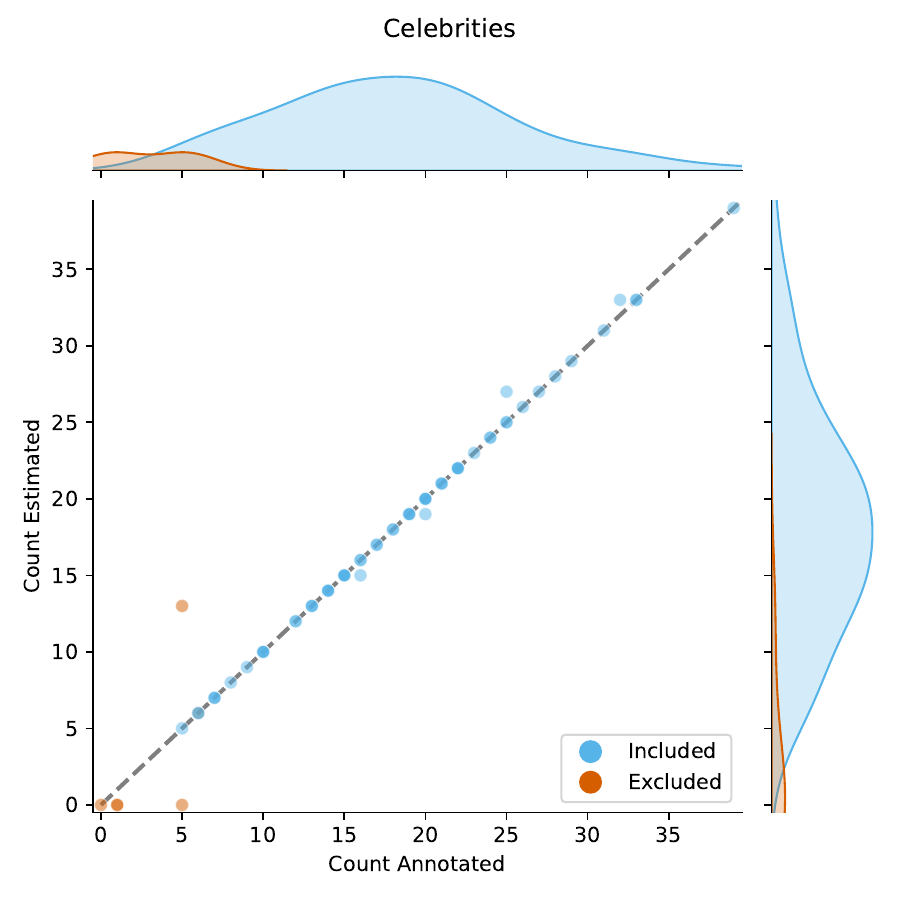}
    \includegraphics[width=0.49\textwidth]{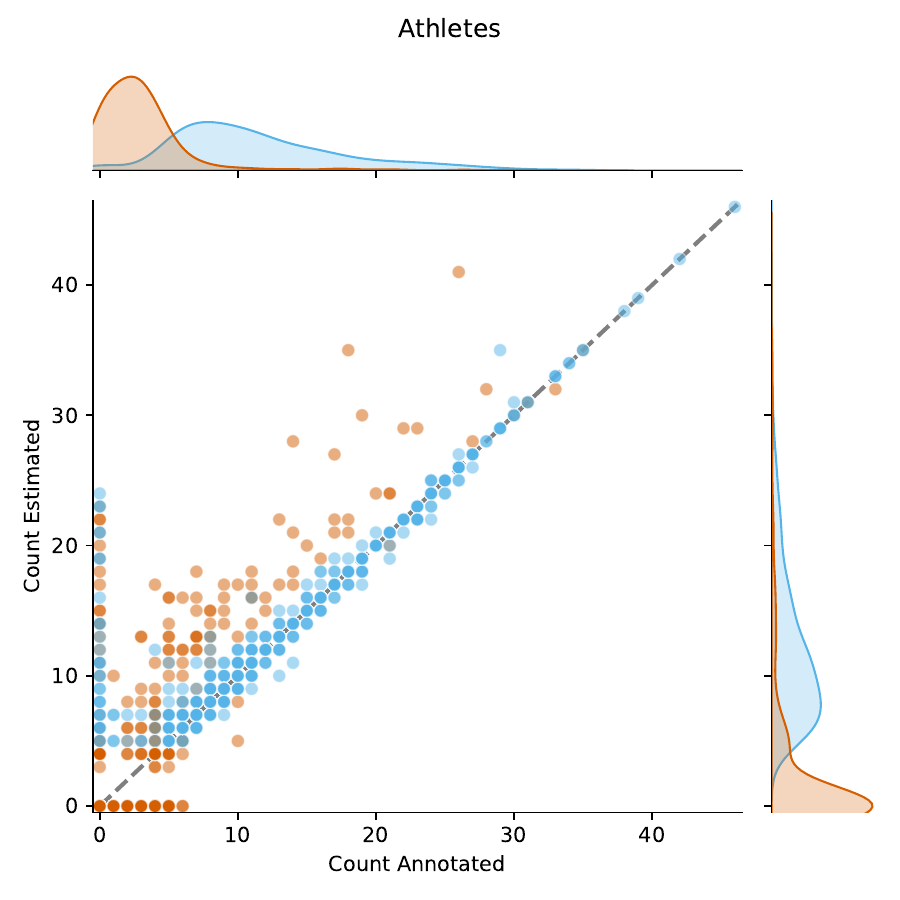}
    \includegraphics[width=0.49\textwidth]{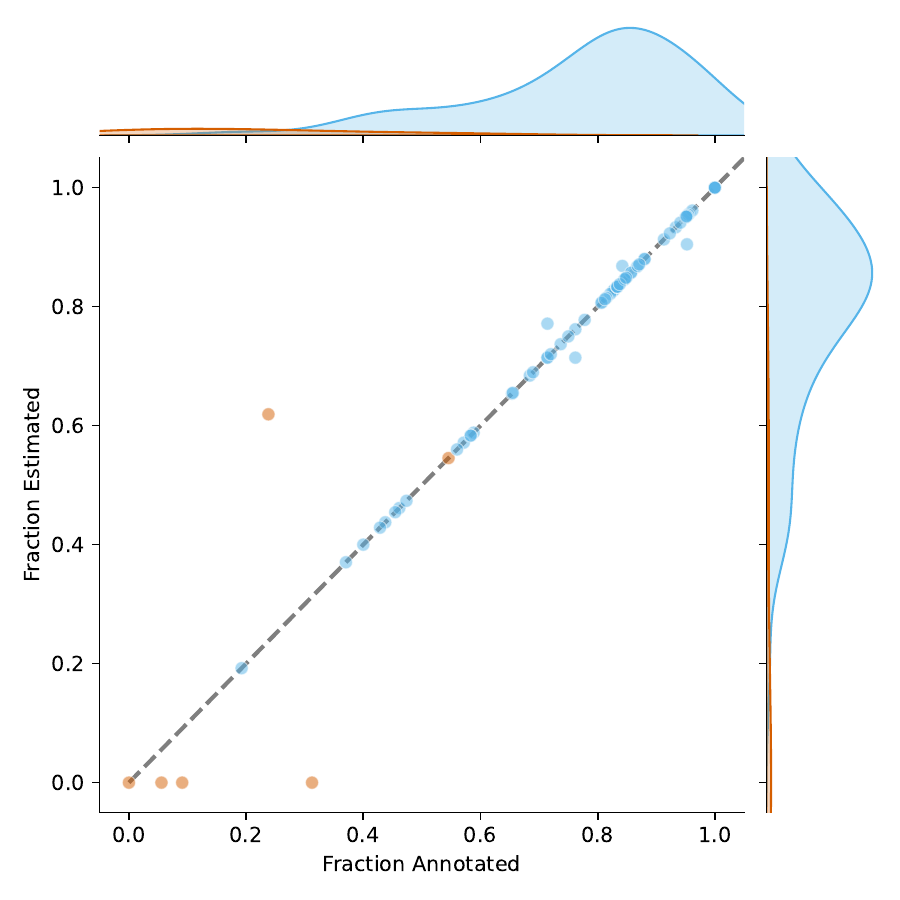}
    \includegraphics[width=0.49\textwidth]{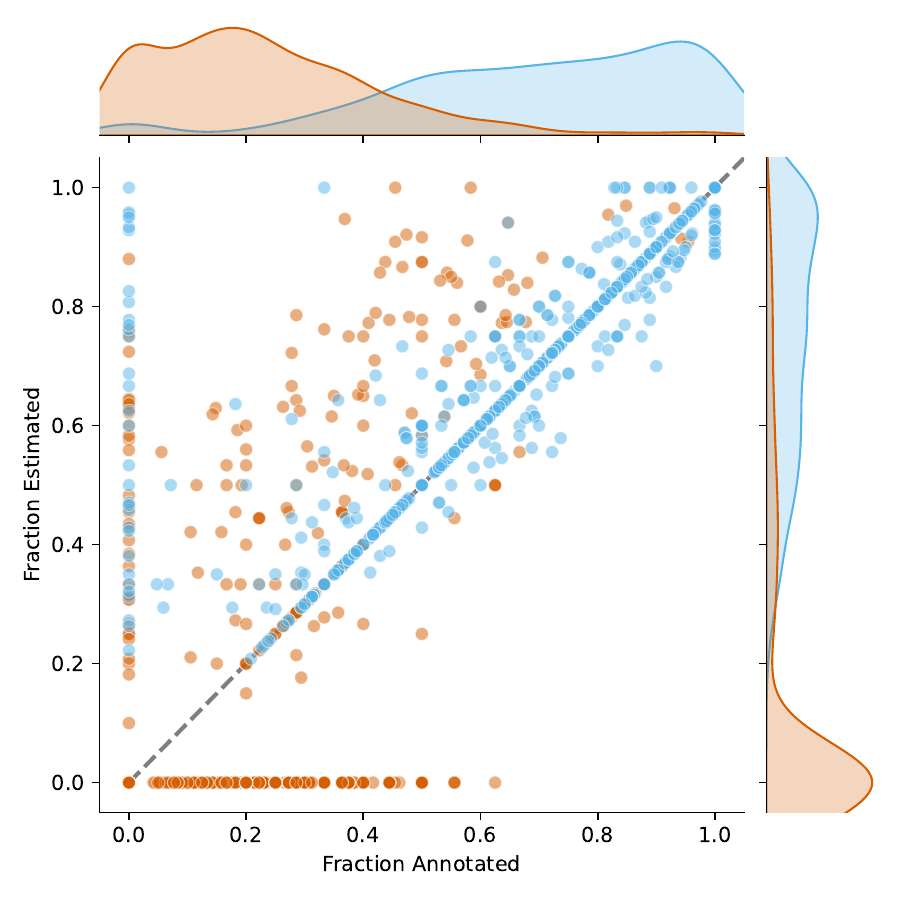}
    
    \caption{{\bf Number (count) and percentage (fraction) of \textit{correct identity} faces per query.} The plots show the absolute numbers (top row) and fraction (bottom row) of correct identity faces for each query $q$ -- one dot per query. The color of each dot shows which queries were excluded from further consideration by our algorithm as described in \cref{sec:gt_estimation}. This plot does not include queries that do not fulfill minimum requirements, which means that all queries contain at least 8 images.}
    \label{fig:correct-ID-scatter-plot}
\end{figure}

\begin{figure}[ht!]
    \centering
\includegraphics[width=.95\textwidth]{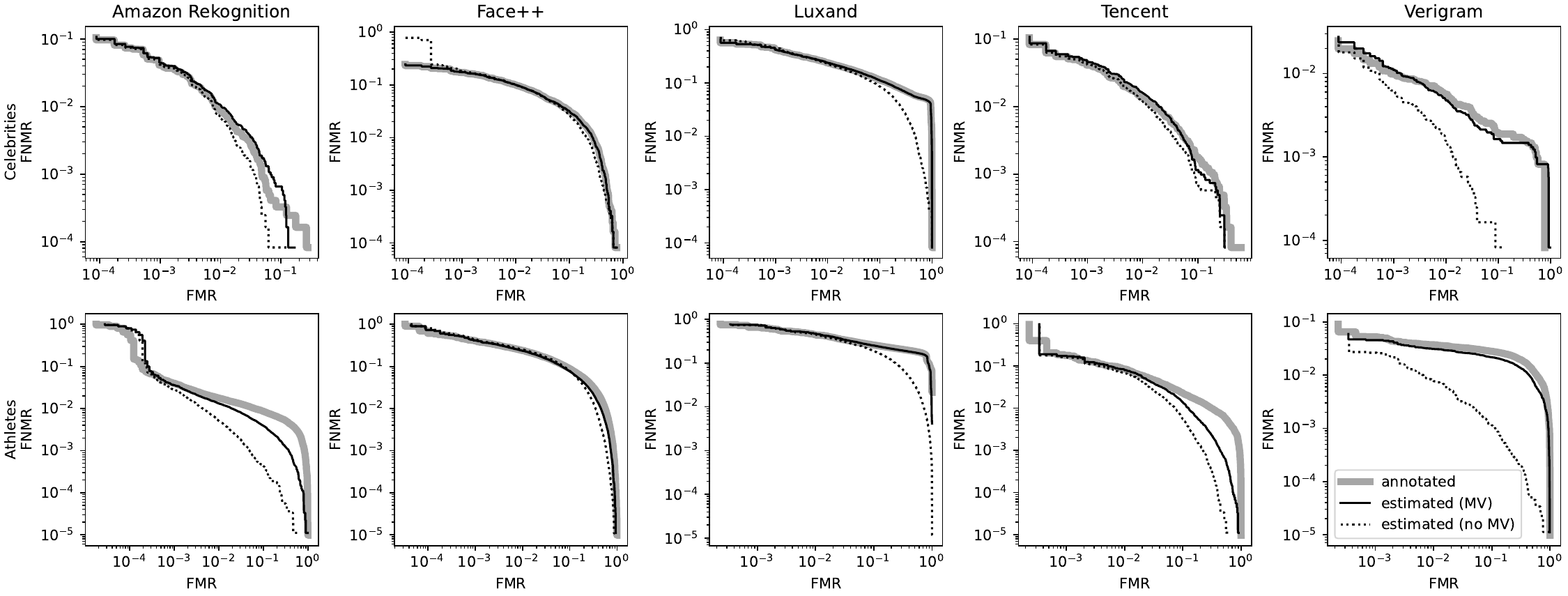}
    \caption{{\bf Effect of majority voting on identity label estimation.} Estimated FMR-vs-FNMR curves are shown before (no MV) and after (MV) consolidation between services. Majority voting yields estimates that are closer to those obtained through hand-annotation. We use majority voting in our method.}
    \label{fig:majority-vote}
\end{figure}

\begin{figure}[ht!]
\centering
\includegraphics[width=\textwidth]{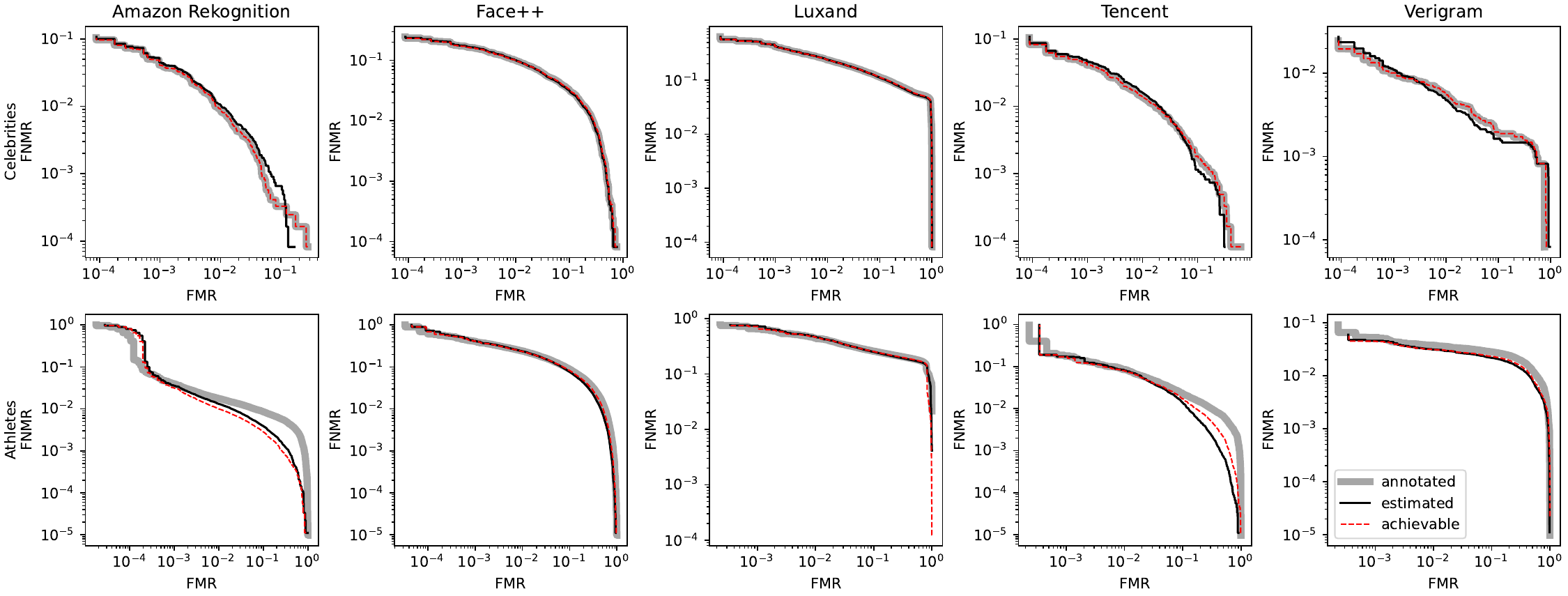}
\caption{{\bf Achievable accuracy for the estimated test set.} The red dashed line indicates the maximum achievable accuracy when $\forall (\hat{y}_i \mid \hat{y}_i \neq -1): \hat{y}_i = y_i$. Compared to the annotated curve, the difference is due to the fact that our method leaves out certain faces where preconditions for correct ID assignment are not given. We can conclude that the error between the annotated and estimated curve for Celebrities mainly stems from wrong assignments ($y_i \neq \hat{y}_i \neq -1$, {\em Type B errors}, see \cref{sec:gt_estimation} for explanation). In contrast, for the Athletes dataset, the visible error is caused by the smaller intersection of annotated and estimated face image sets as we drop significantly more faces in this dataset ({\em Type A error}, see \cref{tab:confusion_matrix}).}
\label{fig:achievable-acc}
\end{figure}

\clearpage
\section{Additional service evaluation results}
\label{appendix:additional-results}

\begin{figure*}[ht!]
    \centering
\includegraphics[width=.825\textwidth]{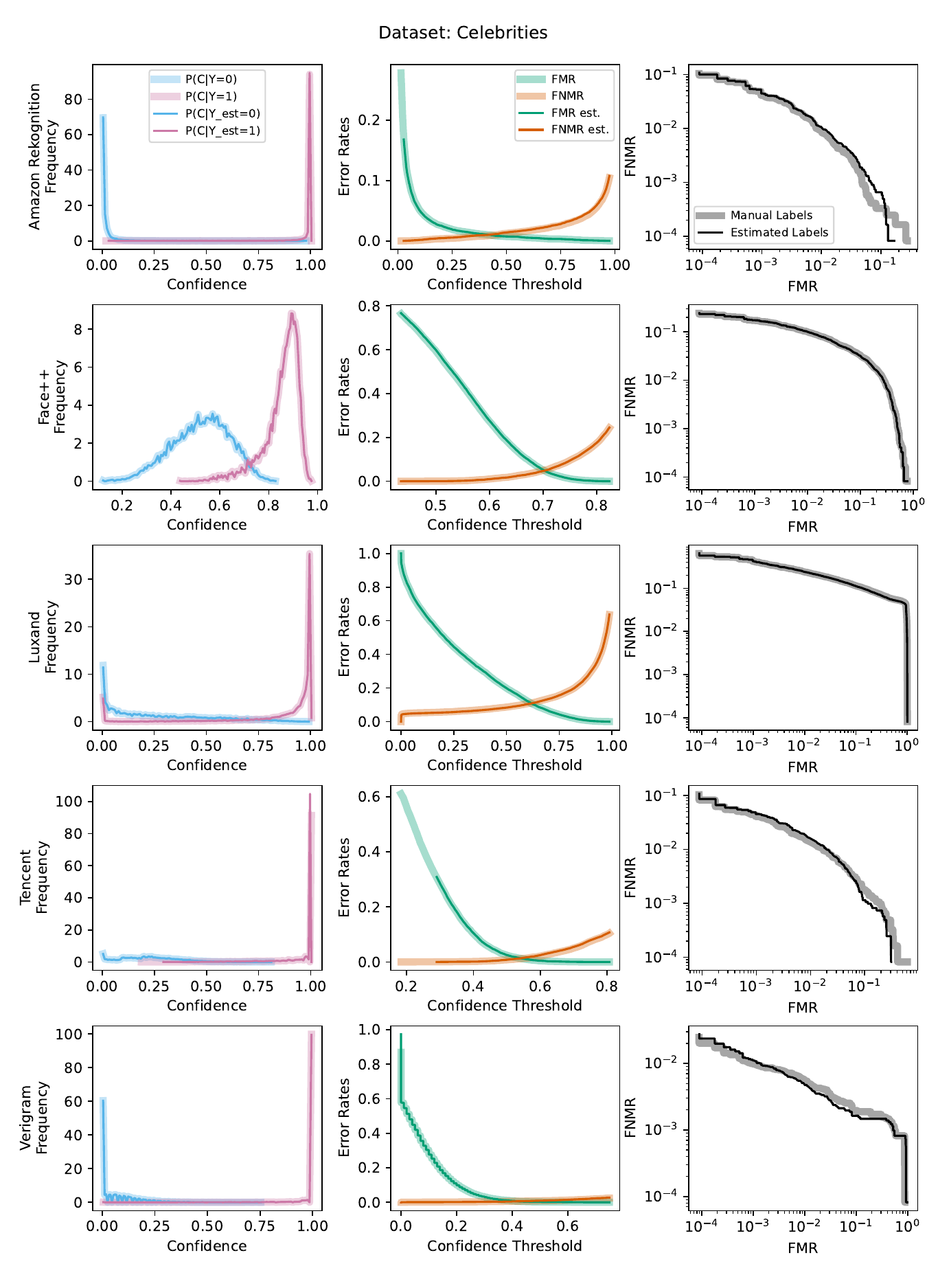}
    \caption{{\bf Verbose service evaluation results for the Celebrities dataset.}}
    \label{fig:results-all-celeb}
\end{figure*}

\begin{figure*}[ht!]
    \centering \includegraphics[width=.825\textwidth]{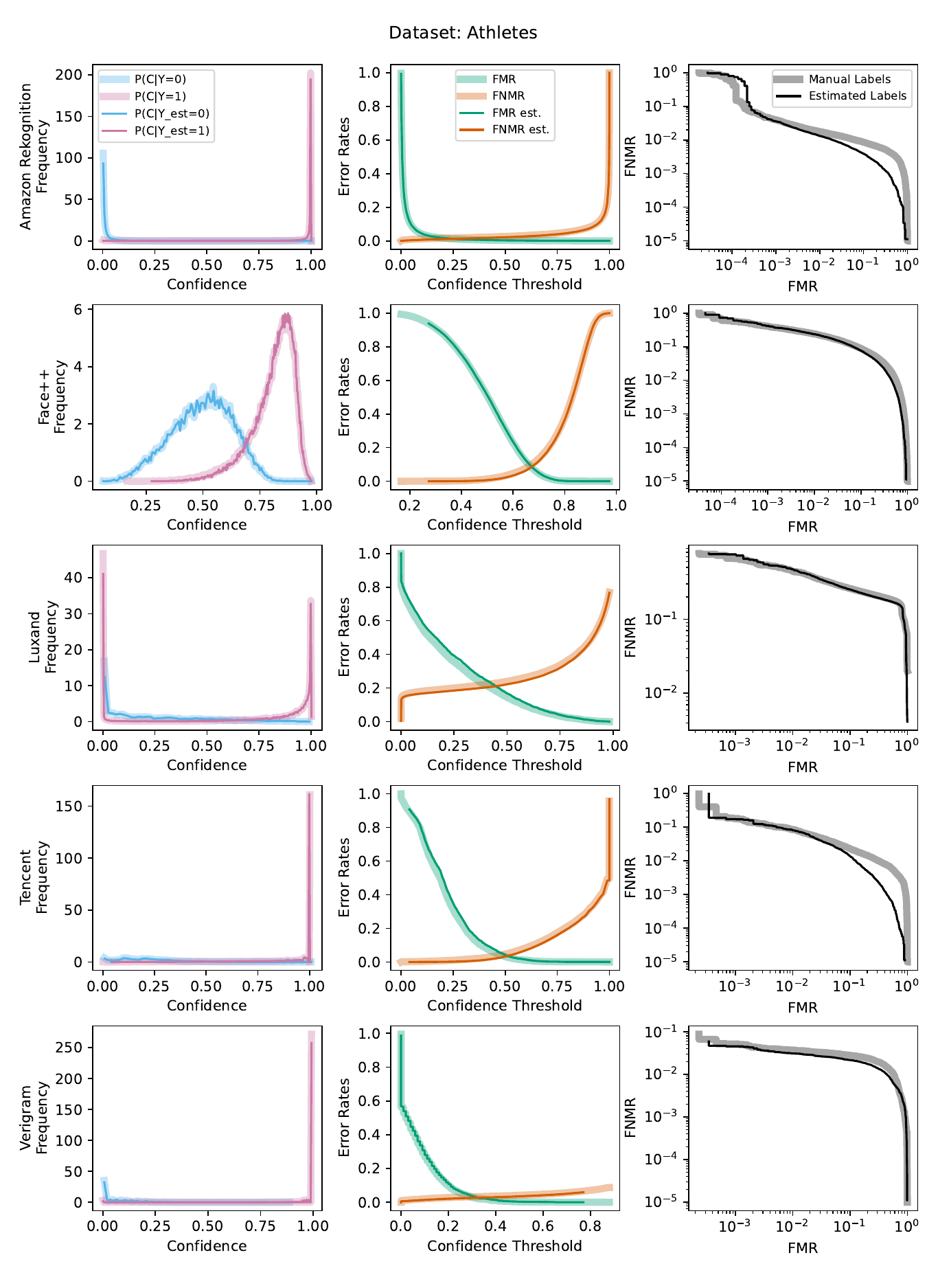}
    \caption{{\bf Verbose service evaluation results for the Athletes dataset.}}
    \label{fig:results-all-athletes}
\end{figure*}

\begin{figure*}[ht!]
    \centering
  \includegraphics[width=\textwidth]{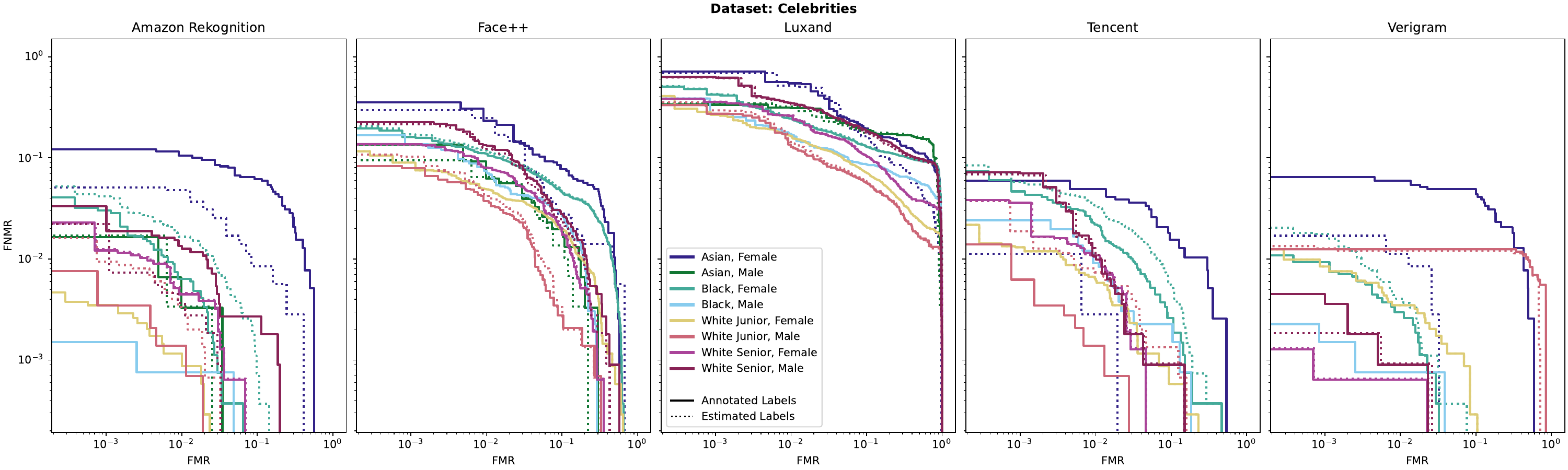}
  \includegraphics[width=\textwidth]{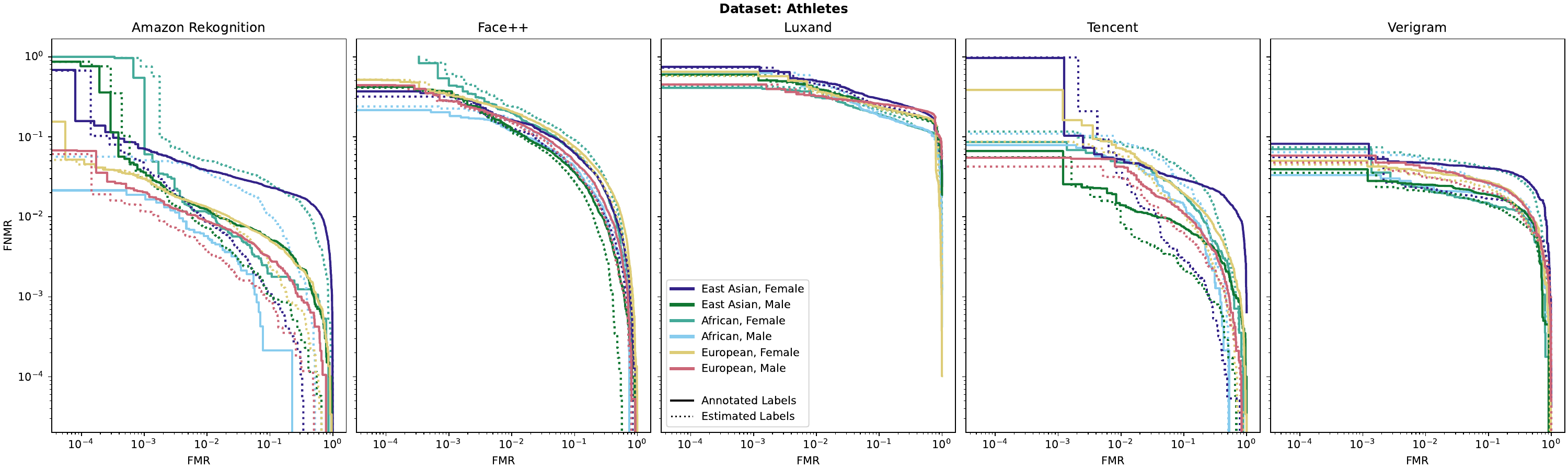}
  \caption{{\bf FMR-FNMR curves by demographic groups.} Each panel shows results for a single service. The top row is based on the celebrities dataset, and the bottom row on the athletes dataset. See also~\cref{fig:bias-equal-error}.
  }
  \label{fig:bias}
\end{figure*}

\begin{figure*}[ht!]
    \centering
  \includegraphics[width=\textwidth]{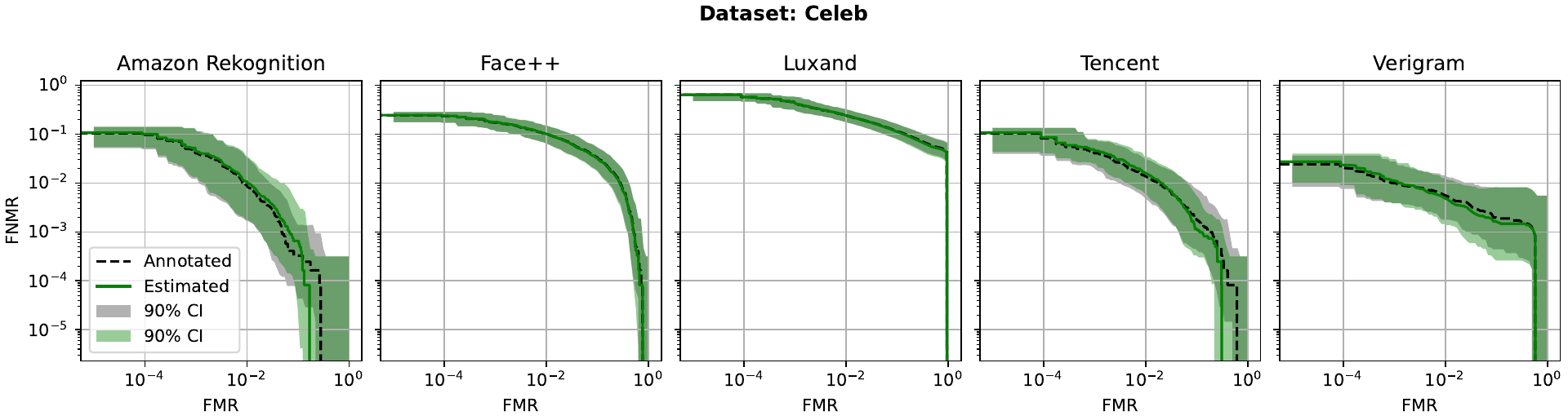}
  \includegraphics[width=\textwidth]{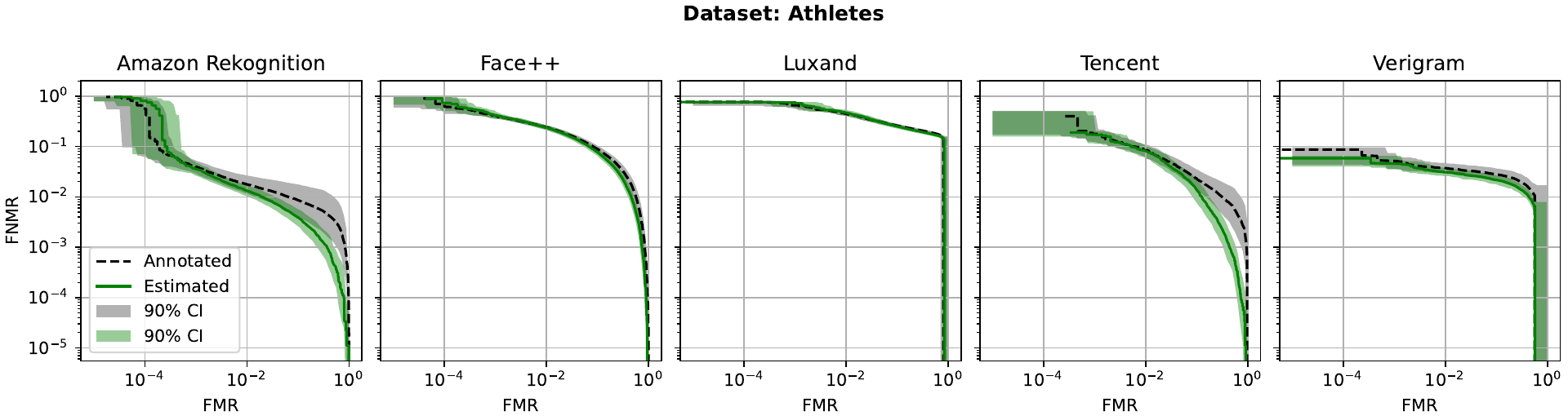}
   \caption{\textbf{FMR-FNMR curves with Wilson confidence intervals \cite{fogliato2024confidence}.} The solid line is our method's estimate, and the dashed line is the hand-labeled annotation. Each panel shows results for a single service for the Celebrities (top row) and Athletes dataset (bottom row).}
  \label{fig:wilson-confidence}
\end{figure*}

\clearpage
\section{Robustness regarding service composition}
\label{sec:service-composition}

\begin{figure}[ht!]
    \centering
        \includegraphics[width=\linewidth]{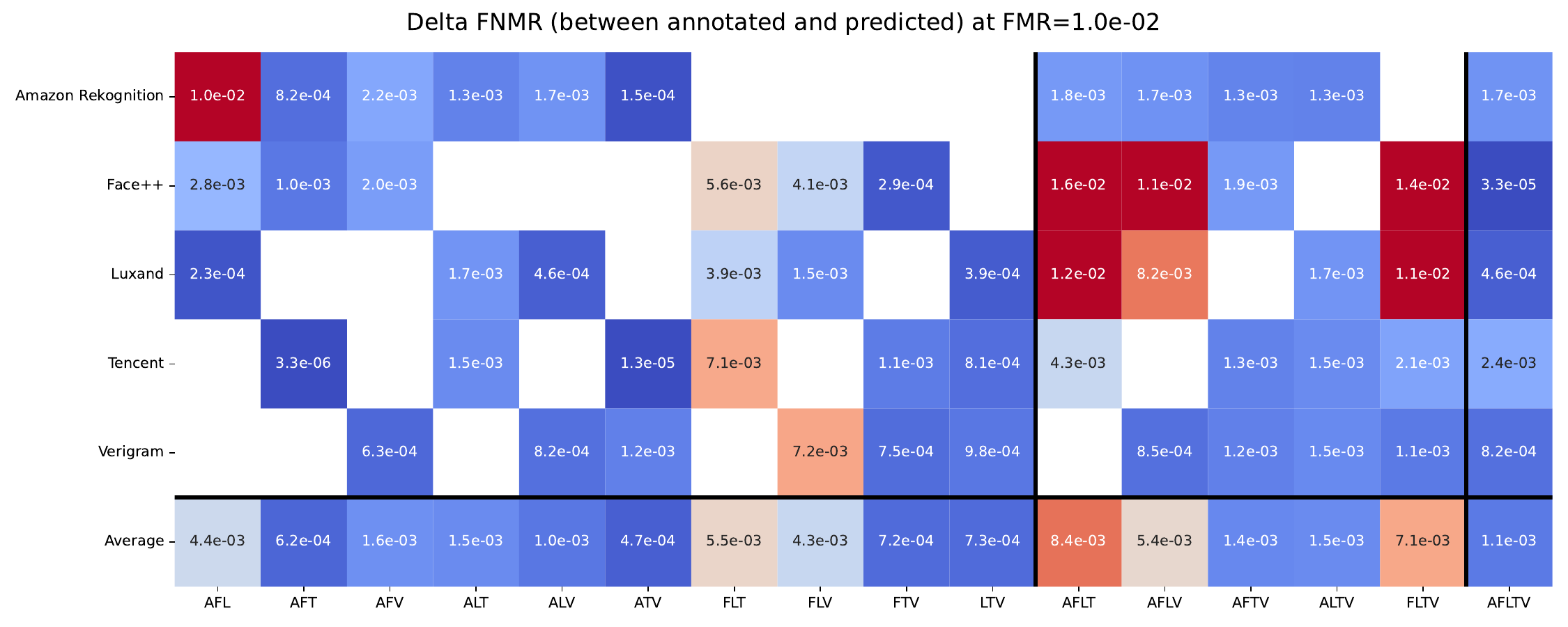}
    \includegraphics[width=\linewidth]{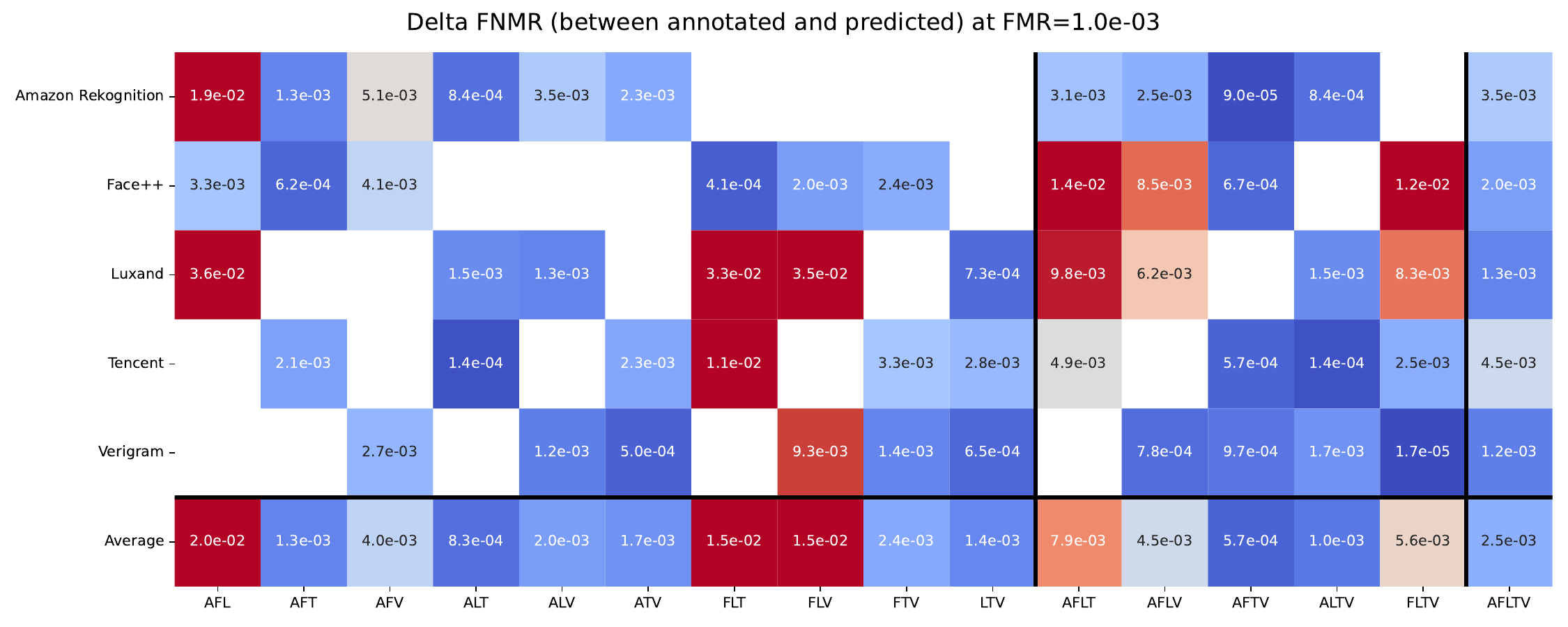}
    \caption{{\bf Effect of service composition on label estimation accuracy using the Celebrities dataset.} We want to test how sensitive our method is w.r.t. the set of included services. To measure the accuracy of our predictions, we calculate \mbox{$\Delta \mathrm{FNMR} = |\mathrm{FNMR}_{estimated} - \mathrm{FNMR}_{annotated}|$} at fixed FMRs of 0.01 (top) or 0.001 (bottom).
    Columns indicate different sets of included services abbreviated with their first letter. White squares indicate that the particular service (row) was not included in this subset (column).
    We test configurations of 3, 4, and 5 included services and find that the inclusion of services that have the lowest accuracy (Luxand) leads to a larger error in many 3-service and 4-service settings, while it can be compensated in the 5-service setting.
    }
    
    \label{fig:service-comp}
\end{figure}

\newpage
\FloatBarrier
\section{Robustness over time}
\label{sec:over-time}

\begin{figure}[ht!]
    \centering
    \includegraphics[width=\linewidth]{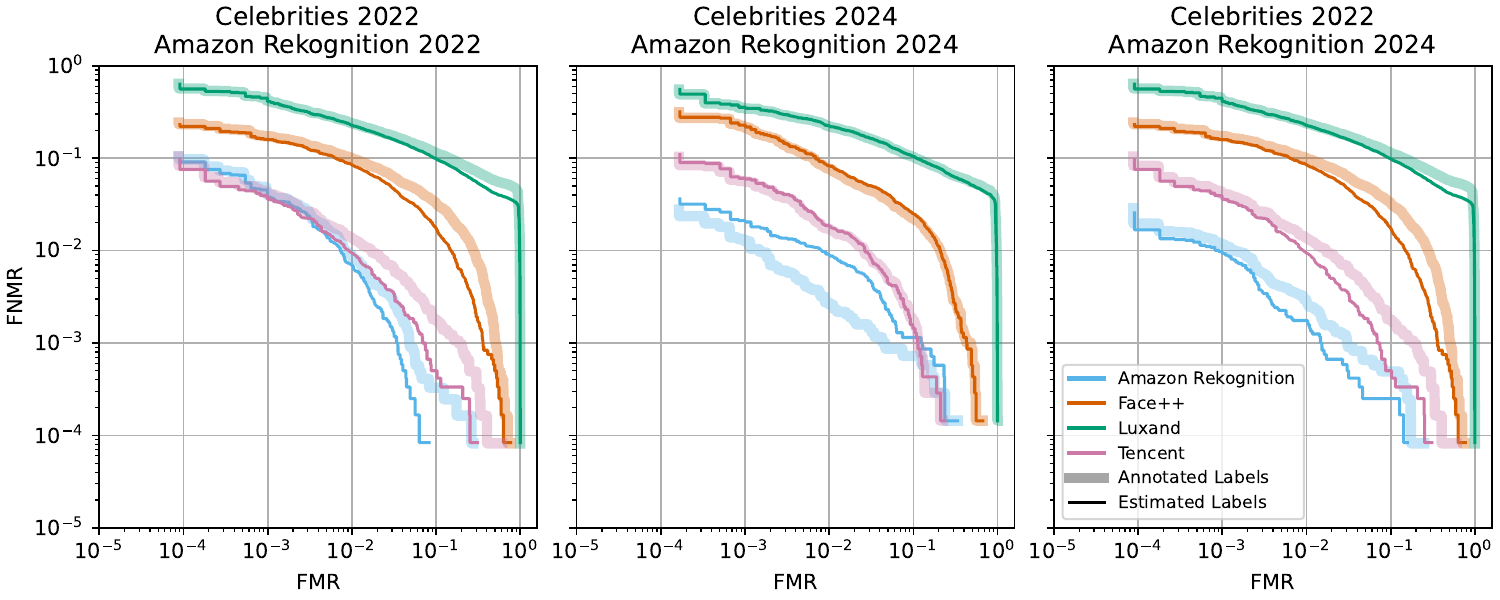}
    \caption{{\bf Probable model update in the Amazon Rekognition service between 2022 and 2024.} In this analysis, we focus on the robustness of our method over time with changing datasets and/or models. The left panel shows estimations based on the Celebrities dataset used in the main paper originally collected in 2022. We did a rerun of our method using the same services and the same list of names in 2024 (mid-panel). As described earlier, this results in a different set of images and possibly a change in the service's underlying model.
    The 2024 rerun shows similar results for three out of four services (Face++, Luxand, Tencent) and improved accuracy for Amazon Rekognition.
    To determine if this improvement is a result of a possible model change, we ran the 2024 version of Amazon Rekognition on the 2022 dataset combined with the other three 2022 services (right panel) and found that the improved service accuracy persists. Therefore, we conclude that a model change has likely happened for the Amazon Rekognition service between 2022 and 2024.
    We show that our method is generally robust over time, even if the underlying evaluation dataset is dynamic by design. Note: Verigram results are omitted as we did not have API access anymore when the 2024 experiments were conducted.
    }
    \label{fig:celeb-aws-old-v-new}
\end{figure}

\newpage
\FloatBarrier
\section{Semi-supervised results}
\label{sec:semi-sup}

\begin{figure}[ht!]
    \centering
    \includegraphics[width=\linewidth]{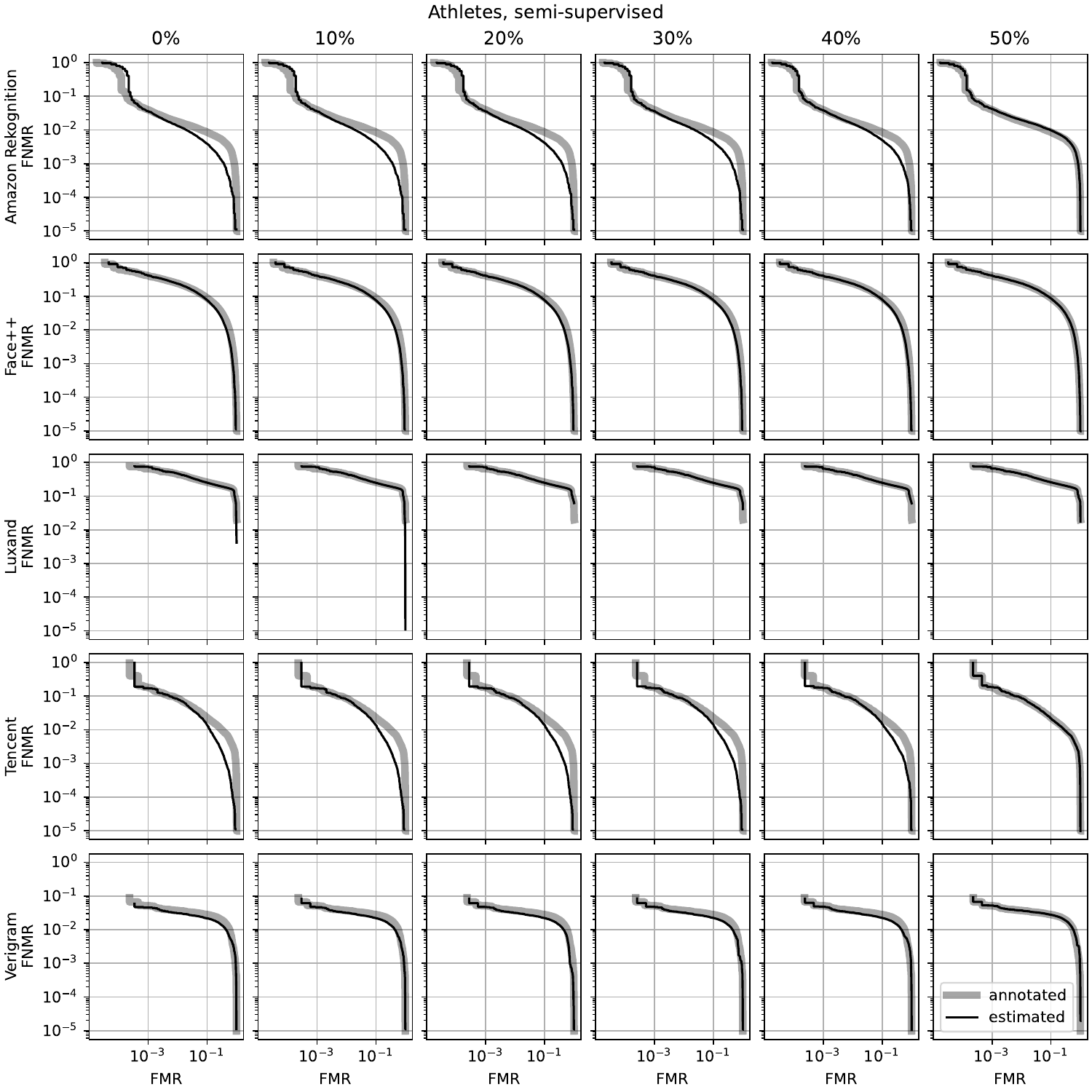}
    \caption{{\bf Semi-supervised FMR-FNMR curves using the Athletes dataset.} Our method allows the combination of estimated and annotated labels for semi-supervised accuracy estimation. Columns indicate the fraction of faces where annotated labels are used.
    As explained in \cref{sec:gt_estimation} and shown in \cref{fig:achievable-acc}, the disagreements between annotated and estimated curves stem from two types of errors. We prioritize to correct {\em Type A} errors by including faces to the estimation dataset that were initially excluded by our method ($\hat{y}=-1$). Once there are no more excluded faces to add, we correct {\em Type B} errors by replacing the estimated labels of those faces that have the highest degree of ambiguity according to their z-values.
    One can see that the curves gradually align with the 100\%-annotated ones and reach near-perfect alignment with 50\%-annotated labels.
    }
    \label{fig:semi-supervised}
\end{figure}

\end{appendices}
\end{document}